\documentclass[10pt,letterpaper]{article}

\usepackage[T1]{fontenc}
\usepackage[utf8]{inputenc}
\usepackage[english]{babel}
\usepackage{mathptmx}
\usepackage[scaled=0.92]{helvet}
\usepackage{amsmath}
\usepackage{amssymb}
\usepackage[letterpaper,top=0.65in,bottom=0.72in,left=0.78in,right=0.78in,
  headheight=14pt]{geometry}
\usepackage[protrusion=true,expansion=true]{microtype}
\usepackage{xspace}
\usepackage[table]{xcolor}
\usepackage{graphicx}
\usepackage{booktabs}
\usepackage{colortbl}
\usepackage{etoolbox}
\usepackage{tabularx}
\usepackage{array}
\usepackage{enumitem}
\usepackage{float}
\usepackage{caption}
\usepackage{fancyhdr}
\usepackage{titlesec}
\usepackage[most]{tcolorbox}
\usepackage[round,authoryear]{natbib}
\usepackage[colorlinks=true,allcolors=argusblue]{hyperref}

\definecolor{sjtuRed}{HTML}{C8161D}
\definecolor{argusblue}{HTML}{315BCE}
\definecolor{argusdeep}{HTML}{24465D}
\definecolor{ink}{HTML}{24465D}
\definecolor{muted}{HTML}{66717D}
\definecolor{papercream}{HTML}{FBF7EE}
\definecolor{papersand}{HTML}{F5E6C8}
\definecolor{papersage}{HTML}{6F9B86}
\definecolor{papergold}{HTML}{D9C58F}
\definecolor{mist}{HTML}{F8F2E7}
\definecolor{rulegray}{HTML}{D8E0E8}
\definecolor{softgray}{HTML}{FFFDF8}

\newcommand{\systemname}{Argus\xspace}

\hypersetup{
  pdftitle={\systemname: A General-Purpose Agentic Reasoning Runtime for Long-Horizon Tasks: Evidence-Governed Progressive Evolution over Durable State},
  pdfauthor={Argus Team},
  pdfsubject={\systemname Technical Report},
  pdfkeywords={objective revision, verification-gated review, long-horizon research, general-purpose agentic runtime, persistent state}
}

\color{ink}
\renewcommand{\arraystretch}{1.10}
\setlist{leftmargin=1.45em,itemsep=1.5pt,topsep=3pt,parsep=0pt}
\AtBeginEnvironment{table}{\rowcolors{2}{papercream}{white}\arrayrulecolor{rulegray}}
\AtEndEnvironment{table}{\rowcolors{2}{white}{white}\arrayrulecolor{black}}

\titleformat{\section}
  {\sffamily\Large\bfseries\color{ink}}
  {\color{argusblue}\thesection}{0.65em}{}
\titleformat{\subsection}
  {\sffamily\large\bfseries\color{ink}}
  {\thesubsection}{0.55em}{}
\titleformat{\subsubsection}
  {\sffamily\normalsize\bfseries\color{ink}}
  {\thesubsubsection}{0.5em}{}
\titlespacing*{\section}{0pt}{1.15em}{0.45em}
\titlespacing*{\subsection}{0pt}{0.85em}{0.28em}
\titlespacing*{\subsubsection}{0pt}{0.65em}{0.2em}

\newcommand{\linkitem}[3]{%
  \raisebox{-0.18\height}{\includegraphics[height=8.5pt]{#1}}\,\href{#2}{#3}}
\fancypagestyle{firstpage}{
  \fancyhf{}
  
  \fancyfoot[C]{%
    \small\sffamily\color{argusblue}
    \linkitem{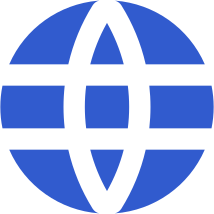}{https://argusbot.cn/}{argusbot.cn}%
    \hspace{0.34in}%
    \linkitem{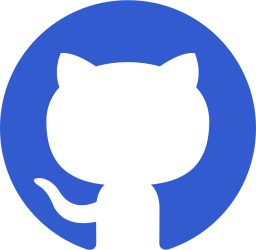}{https://github.com/lbx154/Argus}{github.com/lbx154/Argus}%
    \hspace{0.34in}%
    \linkitem{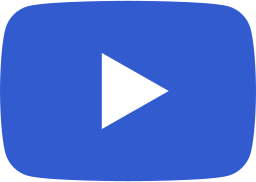}{https://www.youtube.com/watch?v=i8Qy9HCboQE}{youtu.be/i8Qy9HCboQE}%
    \\[3pt]
    \footnotesize\color{muted}\thepage}
}

\newtcolorbox{abstractbox}{
  enhanced,
  colback=papercream,
  colframe=papergold,
  boxrule=0pt,
  arc=1.5pt,
  left=9pt,right=9pt,top=7pt,bottom=7pt,
  before skip=7pt,after skip=6pt
}

\newtcolorbox{paperbox}[1]{
  enhanced,
  colback=softgray,
  colframe=papergold,
  boxrule=0.45pt,
  arc=1.5pt,
  left=8pt,right=8pt,top=6pt,bottom=6pt,
  before skip=6pt,after skip=6pt,
  fonttitle=\sffamily\bfseries\small\color{argusdeep},
  title={#1}
}

\newcommand{\code}[1]{\texttt{\small #1}}

\newcommand{\SWEProDirectAccuracy}{59}
\newcommand{\SWEProArgusAccuracy}{78}

\newcommand{\SWEProTokenRatio}{1.41}

\newcommand{\SWEProTokenReduction}{21}
\newcommand{\SWEProBestTokenReduction}{50}

\newcommand{\SWEProTimeReduction}{15}

\newcommand{\ReviewerTasks}{731}
\newcommand{\ReviewerInvoked}{466}
\newcommand{\ReviewerInvokedPercent}{63.7}
\newcommand{\ReviewerSkipped}{265}
\newcommand{\ReviewerSkippedPercent}{36.3}

\newcommand{\ReviewerFirstBlocked}{35}
\newcommand{\ReviewerRevisionRequested}{43}
\newcommand{\ReviewerVerifierRecovered}{34}
\newcommand{\ReviewerStrictRescues}{22}
\newcommand{\ReviewerTokenRatio}{2.75}
\newcommand{\ReviewerTimeRatio}{1.80}

\newcommand{\PaperCasePapers}{6}
\newcommand{\PaperCaseCompleted}{6}
\newcommand{\PaperCaseCampaignHours}{640}
\newcommand{\PaperCaseMissions}{254}
\newcommand{\PaperCaseRounds}{576}
\newcommand{\PaperCaseContinues}{286}
\newcommand{\PaperCaseSessionRolls}{89}
\newcommand{\PaperCaseRollbacks}{16}
\newcommand{\PaperCaseReviewSnapshots}{436}

\begin{document}
\thispagestyle{firstpage}

\vspace{0.04in}

\noindent{\sffamily\small\bfseries\color{muted}
  \MakeUppercase{\systemname} TECHNICAL REPORT \enspace\textperiodcentered\enspace AUG 2026}\par
\vspace{0.085in}

\noindent{\raggedright\sffamily\fontsize{23}{26}\selectfont\bfseries\color{ink}
  \systemname: A General-Purpose Agentic Reasoning Runtime for Long-Horizon Tasks}\par
\vspace{0.055in}
\noindent{\raggedright\sffamily\fontsize{13}{16}\selectfont\bfseries\color{argusdeep}
  Evidence-Governed Progressive Evolution over Durable State}\par
\vspace{0.06in}
\noindent{\raggedright\sffamily\normalsize\bfseries\color{ink}
  \systemname Team}\par
\vspace{0.085in}

\noindent\hbox to\linewidth{%
  \raisebox{-0.5\height}{\includegraphics[height=0.227in]{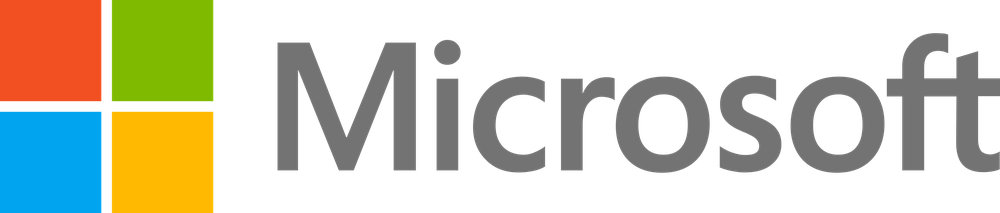}}\hfill
  \raisebox{-0.5\height}{\includegraphics[height=0.30in]{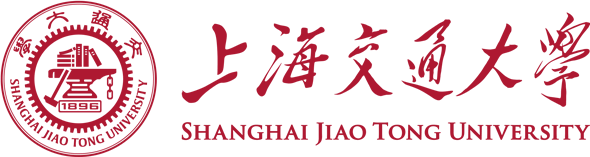}}\hfill
  \raisebox{-0.5\height}{\includegraphics[height=0.30in]{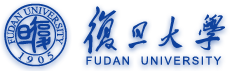}}\hfill
  \raisebox{-0.5\height}{\includegraphics[height=0.30in]{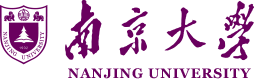}}\hfill
  \raisebox{-0.5\height}{\includegraphics[height=0.30in]{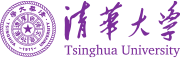}}%
}
\vspace{0.07in}

\noindent\hbox to\linewidth{%
  \raisebox{-0.5\height}{\includegraphics[height=0.227in]{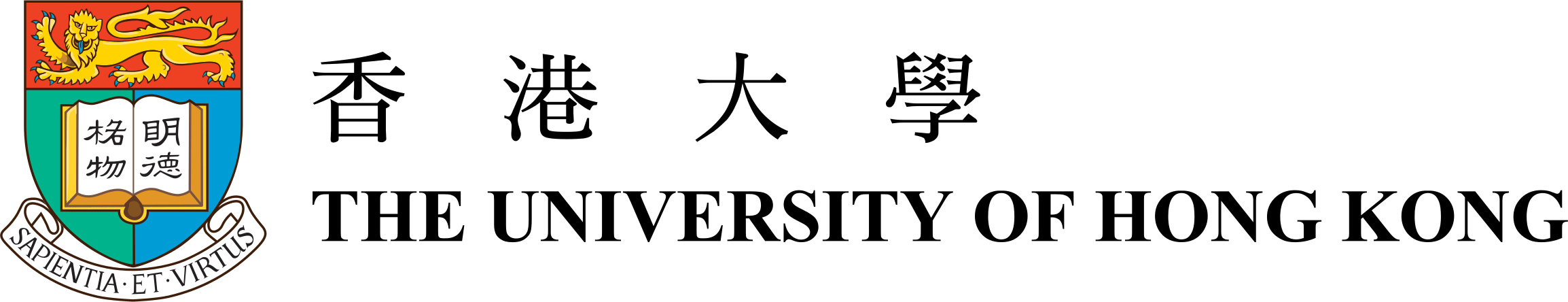}}\hfill
  \raisebox{-0.5\height}{\includegraphics[height=0.30in]{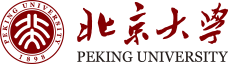}}\hfill
  \raisebox{-0.5\height}{\includegraphics[height=0.196in]{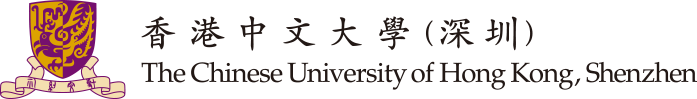}}\hfill
  \raisebox{-0.5\height}{\includegraphics[height=0.30in]{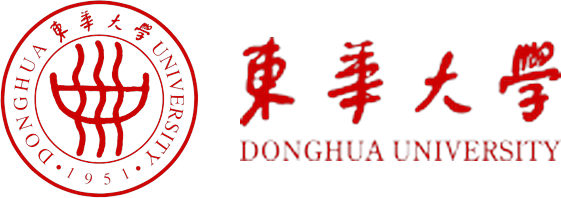}}%
}
\vspace{0.10in}

\begin{abstractbox}
\small
\textbf{Abstract.}
Recursive self-improvement currently works where the supervision signal is dense.
Kernel optimization, competitive benchmarks, and any task with a fast numeric score
give an agent something to climb, and agents that climb them improve. The tasks that
matter most in research supply no such signal: the objective is underdefined at the
start, the right measurement is itself part of the problem, and the only available
feedback is sparse, delayed, and contested. A system tuned to a dense score is a
student who can only answer exam questions. We present \systemname, a persistent
agentic runtime built for the other case, in which an expert and the system
\emph{co-evolve} a problem statement: the runtime carries a campaign long enough to
turn an underdefined goal into a defined one, and surfaces the specific contradiction
that requires a human decision instead of guessing past it. Manager, Planner,
Engineer, and Reviewer execute bounded missions over durable project state, and the
working contract separates stable user intent from the operational objective,
constraints, and verification criteria that evidence may refine. Verification-gated
admission makes refinement cumulative rather than ad hoc: candidate memories, skills,
procedures, verifiers, routing decisions, and rejected routes become reusable only
after role-owned review and, where available, task-native verifier evidence. Model
weights remain fixed, so self-evolution occurs in the persistent runtime state and
control policy, and ordinary rounds run unattended between operator-owned escalation
points.

In benchmark runs with autonomous mission execution, the same harness remains effective across seven
GPT-5.5 benchmark arenas. It reaches approximately \SWEProArgusAccuracy\% on
SWE-Bench Pro versus \SWEProDirectAccuracy\% for Direct Copilot, while using
\SWEProTokenRatio$\times$ the aggregate Tokens. After verification-gated runtime
self-evolution, mature SWE-Bench Waves use \SWEProTokenReduction\% fewer solve input
Tokens and \SWEProTimeReduction\% less active workflow time per task than startup
Waves. This observational trajectory is not monotone; it also records
\ReviewerVerifierRecovered{} verifier recoveries and \ReviewerStrictRescues{}
strict review-loop rescues. Across the broader suite, \systemname reaches 76.8\% on
AARRI-Bench research tasks and a 28.0 gap on mathematical data synthesis, alongside
competitive GPU-kernel and language-model-training results. Its artifacts extend
beyond benchmark endpoints: an optimized RWKV6 kernel was merged into the upstream
Flash Linear Attention repository. A multi-day mathematical campaign retains one
falsified route and six proof-backed frontier updates; six paper pipelines with
autonomous mission execution
span \PaperCaseMissions{} missions and \PaperCaseRollbacks{} Stage rollbacks before
reaching submission completion. Two verticals close against external checkers the
runtime does not own: an inference accelerator certified for its demonstrated scope,
and a materials campaign in which the admitted method is simpler than the published
one it replaces. These results show a general, self-evolving harness
that can revise, recover, and accumulate verified approaches without forcing every
long-horizon trajectory into a success claim. The retained trajectories also form
structured data for future supervised and reinforcement learning.
\end{abstractbox}

\noindent{\footnotesize\color{muted}
  \par}

\begin{figure}[!t]
  \centering
  \includegraphics[width=\linewidth]{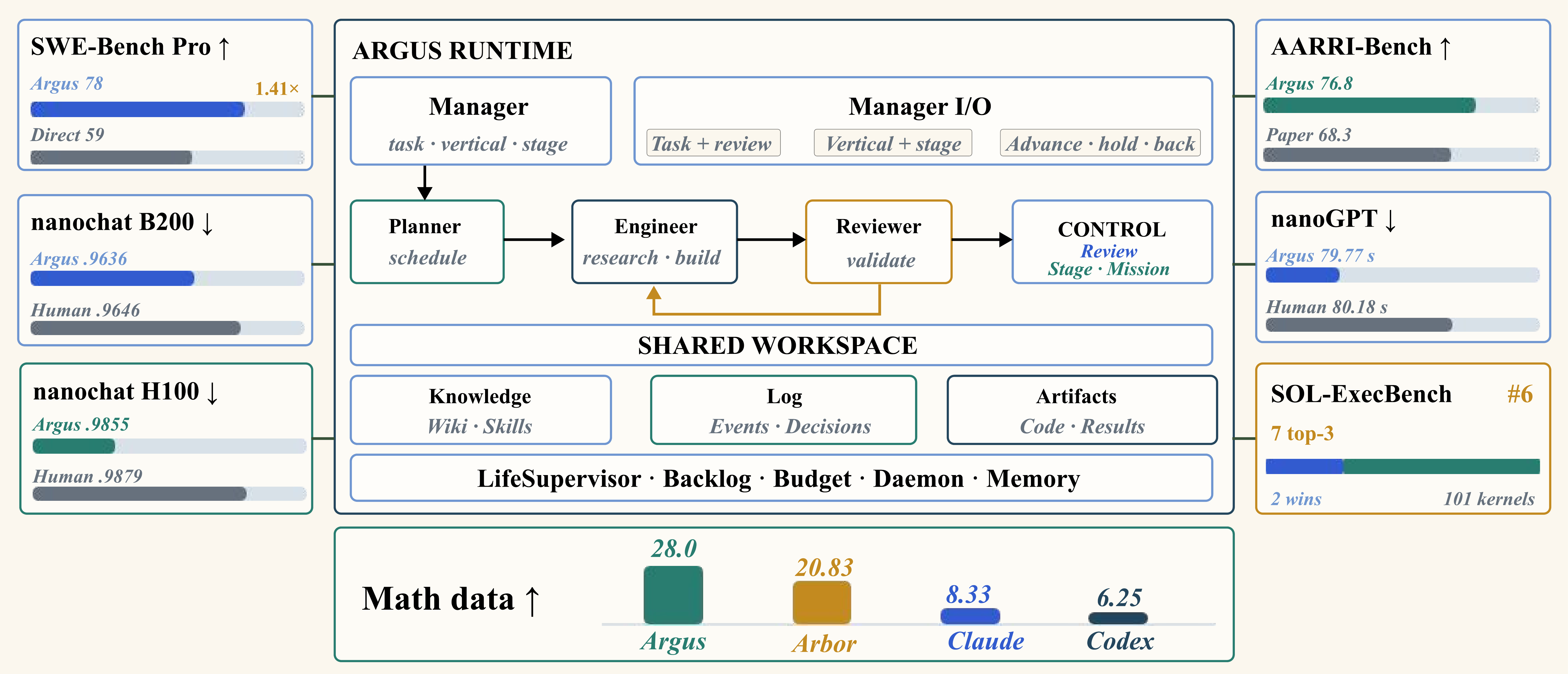}
  \caption{\systemname runtime and breadth evidence. The central diagram shows
  Manager authority over tasks, research verticals, and Stage transitions;
  Planner, Engineer, and Reviewer operate over a shared workspace whose knowledge,
  event log, artifacts, backlog, budget, daemon, and memory persist across bounded
  missions. The surrounding cards report outcomes from seven task-native
  evaluations. Their bars use independent scales and indicate comparison direction
  with arrows; they are breadth evidence rather than a single normalized
  leaderboard.}
  \label{fig:overview}
\end{figure}

\section{Introduction}
\label{sec:introduction}

Self-improving agents work today wherever the supervision signal is dense. Give a
system a kernel to make faster, a leaderboard to climb, or a unit test to turn green,
and iteration has something to optimize: every attempt returns a number, and the
number says whether to keep going. Progress under those conditions is real, and it is
also narrow. It describes a student who has become excellent at answering questions
someone else has already written down.

The tasks that motivate this report supply no such signal. A mathematical campaign
rarely terminates in the theorem it set out to prove: intermediate bounds,
counterexamples, and reformulations are the normal output, and they carry real value.
A software request is often underspecified until a candidate implementation exposes
what was missing. In system verification, the specification and the implementation may
both be wrong, and deciding which one to change is itself part of the work. In chip
design and in materials research, the measurement that would settle the question is
frequently the thing under construction. What these share is not difficulty but
\emph{underdefinition}: at the moment work starts, nobody, human or machine, can state
the objective precisely enough to optimize against it.

This changes what a runtime is for. If the score cannot be trusted to steer, the
system's job is not to climb it but to hold a campaign together long enough for the
objective itself to become clear, and to route the specific contradictions it cannot
resolve to the human expert who can. We call this \emph{expert--agent co-evolution}:
the standing intent stays stable while the operational objective, constraints, and
verification criteria are refined by evidence produced along the way, and the moments
that require human authority are surfaced explicitly rather than guessed past.

Permitting that revision is difficult because it is externally indistinguishable from
failure. A system that abandons its stated target may have discovered that the target
was misspecified, or it may have failed and then rationalized. Nothing in the final
artifact separates the two. A runtime that allows objective revision without resolving
this ambiguity invites \emph{goal drift}: the objective degrades toward whatever the
executor happens to be able to finish. The standard response is to remove the choice,
fixing the objective and optimizing execution toward it.

Prior work follows that response. ReAct, SWE-agent, and OpenHands established
practical interfaces between models and external environments
~\citep{yao2023react,yang2024sweagent,wang2024openhands}. The AI Scientist,
CycleResearcher, AutoSci, FARS, and Arbor extend this interaction into iterative
research programs covering planning, experiments, review, memory, and scientific
communication~\citep{lu2024aiscientist,weng2025cycleresearcher,qian2026autosci,
tang2026fars,jin2026arbor}. These systems differ in how they search, remember, and
review, but they share the assumption that the objective is given. Their reported
progress makes the remaining problem visible: agent success declines as software-task
horizons grow~\citep{kwa2025longtasks}, and the horizons that matter most are those
over which the objective is least likely to survive unchanged.

We take the opposite position: pivoting should be permitted and made safe by
verification. An admissible pivot is supported by evidence that the previous route
or objective was unreachable or misspecified, admitted through an explicit role
boundary, and recorded so that later missions inherit both the change and its
justification. Verification is therefore not a quality filter applied after
execution. It is the mechanism that makes pivoting distinguishable from drift.

Three properties must hold before any such certification is possible, and each fails
in long-running agents. \emph{Continuity} fails when a growing transcript is compacted
or dropped, so the evidence that motivated a revision no longer exists when the
revision is proposed. \emph{Acceptance} fails when the component that performs the
work also declares it complete, so a revision is certified by the party with the
strongest incentive to accept it. \emph{Experience} fails when only the final artifact
survives, so refuted routes cannot later be cited as evidence that an objective is
unreachable. Hierarchical and workflow-oriented memory systems address parts of this
problem~\citep{packer2023memgpt,wang2024workflowmemory,xu2025amem}. A larger context
window extends session continuity; certifying a change of objective additionally
requires explicit state, ownership, and update rules.

We present \systemname (Autonomous Research Generation and Understanding System), a
general-purpose agentic runtime for long-horizon reasoning. Its central abstraction
is a sequence of \emph{bounded missions} executed against a durable project state.
Four model-driven roles divide authority: a Manager anchors the objective and
campaign state, a Planner selects the next units of work, an Engineer implements
and evaluates them, and a Reviewer inspects artifacts and run outputs before issuing
the completion verdict when independent review is required or requested. Allowed
low-risk tasks may instead use recorded Engineer self-review. These roles operate across three planes---control,
execution, and records---so that scheduling, work, and record keeping remain
separable even though they participate in one continuous loop.

At the report level, we summarize the evolving working contract as
$K_t=(\iota,o_t,c_t,v_t)$. Here $\iota$ is the standing user intent that the campaign
must preserve; $o_t$ is the current operational objective; $c_t$ is the set of known
constraints; and $v_t$ specifies the verification criteria at mission $t$. We use
$X_t$ for user-visible clarifications, priorities, and unresolved questions. This
notation separates an evidence-backed refinement of the operational contract from a
silent change of intent. Section~\ref{sec:problem} formalizes how evidence, recorded
verdicts, and the required user or Manager authority admit a material update.

Deployment provides qualitative motivation for this design. Expert users of \systemname
report that a substantial part of its practical value lies in where it stops: the
system declines to continue on an underspecified objective, and that refusal surfaces
constraints the user had not stated. Related internal system-verification cases
involve real researchers and project-specific details that cannot be disclosed in
this report; we therefore record the observed use as qualitative motivation rather
than a public prospective user-study result. After initial assignment, the current runtime can execute
ordinary rounds unattended and pause explicitly for operator-owned decisions; the
public traces do not provide a measured zero-touch rate.

A pivot should not reset the campaign. The runtime self-evolves the approach between
the standing intent and the current operational contract: admitted Stages retain
attempted routes, measurements, verdicts, accepted and rejected results, reusable
skills, verifiers, and routing decisions. Candidate updates enter persistent state
only after the authorized role checks the relevant artifacts and task-native evidence;
the gate may include an official verifier, independent Reviewer, or explicitly allowed
Engineer self-review according to task policy. Model parameters remain fixed, but
later missions therefore begin from a changed search policy rather than merely a
longer transcript. We call this mechanism \emph{verification-gated fixed-model runtime
self-evolution}. Section~\ref{sec:results} reports its longitudinal evidence as
observational rather than a controlled learning ablation.

We sharpen this design language into a process-to-capability theory. The final
artifact is a projection of a richer typed trajectory; reviewed compression retains
the parts needed for later decisions; and an accepted update counts as compounding
only when it reduces future task risk or resource cost. This separates three claims
that are often conflated: more process information is available, a review gate
improves admission quality, and retained state actually helps later work.

Overall, the evaluation asks whether one self-evolving runtime can combine broad
task capability, verified accumulation across missions, and sustained research
delivery. The evidence is organized in three parts. First, seven benchmark arenas
cover software repair, GPU kernels, language-model training, training speed,
research-assistant tasks, and mathematical data synthesis. Beyond SWE-Bench Pro,
the suite includes 76.8\% on AARRI-Bench and a 28.0 mathematical-data gap, both above
their reported references; a de-duplicated inventory separately records 41 research
artifacts across six programs. Second, the 731-task SWE-Bench Pro trajectory supports
longitudinal analyses of Skill/Wiki evolution, verifier recovery, Reviewer
intervention, and mature-stage operating cost. Third, vertical evidence tests whether
retained state changes later research: one mathematical campaign preserves a
falsified route and six proof-backed theorem-frontier updates, while six
paper-production campaigns span \PaperCaseMissions{} missions and
\PaperCaseRollbacks{} Stage rollbacks before producing final AAAI- and ACL-formatted
manuscripts. Two further verticals carry the same discipline into domains with
unforgiving external checkers: a chip certified for its demonstrated scope through
mapped synthesis and static timing, and a materials campaign that replaces a
published sampling method with a simpler one after removing a confound in the
original comparison.

This report makes seven contributions:

\begin{enumerate}[label=\textbf{C\arabic*.}]
  \item We formulate long-horizon research as verified pivoting over a compact
    report-level contract $K_t=(\iota,o_t,c_t,v_t)$ and explicit user decision state
    $X_t$, with each variable defined above. The analytical \code{ManagerAdmit}
    operator makes evidence, authority, and
    provenance requirements explicit without claiming one atomic implementation API.
  \item We present a runtime in which contract refinement is represented through
    typed, logged, role-owned surfaces rather than one atomic operation or an implicit
    consequence of replanning. \code{GoalContract} revisions, Manager stage
    transitions, no-go rollbacks, and recorded verdict sources give material changes
    an author, a precondition, and an audit trail. The same admission discipline
    gates fixed-model runtime self-evolution: generated memories, skills, verifiers,
    routing decisions, and rejected routes become reusable only after evidence checks
    and an authorized commit.
  \item We measure where the runtime spends verification and where it refuses to
    terminate. Of \ReviewerTasks{} SWE-Bench Pro tasks, \ReviewerInvoked{} invoke an
    independent Reviewer and \ReviewerSkipped{} use Engineer self-review. The Reviewer
    withholds completion on \ReviewerRevisionRequested{} tasks, of which
    \ReviewerVerifierRecovered{} later pass the official verifier and
    \ReviewerStrictRescues{} complete the strict review loop; a further
    \ReviewerFirstBlocked{} tasks are declared blocked rather than reported complete.
    A longitudinal view of the same run shows mature Waves using
    \SWEProTokenReduction\% fewer solve input Tokens and \SWEProTimeReduction\% less
    active workflow time per task than startup.
  \item We present a six-project autonomous paper-production case study covering
    \PaperCaseCampaignHours{} campaign-hours, \PaperCaseMissions{} bounded missions,
    \PaperCaseRounds{} Engineer rounds, \PaperCaseContinues{} Reviewer revisions,
    \PaperCaseSessionRolls{} session rolls, and \PaperCaseRollbacks{} Stage rollbacks,
    with all six canonical pipelines reaching submission completion. A representative
    campaign uses seven no-go rollbacks to reframe a failed method search as a
    4,500-row negative-results study, then repairs two late submission defects
    without resetting the research state.
  \item We reconstruct one mathematical campaign as a role-resolved Agent trace,
    making the Manager, Planner, Engineer, and Reviewer actions visible across
    multiple bounded missions, and retaining one falsified route alongside six
    accepted theorem-frontier updates.
  \item We report a capability floor across seven benchmark arenas in one unified
    table that preserves each benchmark's backbone, backend, protocol, native metric,
    and reference, establishing that the verification machinery does not cost
    end-task performance: on SWE-Bench Pro, \systemname reaches approximately
    \SWEProArgusAccuracy\% accuracy versus \SWEProDirectAccuracy\% for Direct Copilot
    at \SWEProTokenRatio$\times$ aggregate Tokens.
  \item We report external adoption of an \systemname-produced artifact: a TileLang
    \code{RWKV6} kernel reviewed by an outside maintainer and merged into
    \code{fla-org:main}.
\end{enumerate}

The remainder of the report follows a conventional research-paper organization.
Section~\ref{sec:related} positions the system against prior agent and autonomous
research work. Section~\ref{sec:problem} formalizes the operating problem;
Section~\ref{sec:method} presents the runtime; Sections~\ref{sec:methodology} and
\ref{sec:results} describe the evaluation protocol and results; and
Sections~\ref{sec:discussion}--\ref{sec:conclusion} discuss interpretation,
limitations, and conclusions. The appendix contains detailed experimental tables
and notation.

\section{Related Work}
\label{sec:related}

\subsection{Autonomous Research Systems}

Autonomous research systems range from tightly scoped experiment loops to complete
research lifecycles. Karpathy's \emph{autoresearch} repeatedly modifies, trains, and
evaluates a model under a fixed compute budget~\citep{karpathy_autoresearch}; AIDE
organizes iterative code search for machine-learning improvements
~\citep{jiang2025aide}. FunSearch and AlphaEvolve extend evaluator-guided program
search to mathematical, algorithmic, and systems problems
~\citep{romera2024funsearch,novikov2025alphaevolve}. These systems show how a clear
objective and an executable evaluator can turn model calls into sustained search.

Full-lifecycle systems add problem selection, planning, experimentation, review,
and scientific communication. The AI Scientist introduced an end-to-end research
pipeline, while AI Scientist-v2 added agentic tree search and produced a
workshop-accepted paper~\citep{lu2024aiscientist,yamada2025aiscientistv2}. Agent
Laboratory organizes specialized roles across the research process
~\citep{schmidgall2025agentlab}, and CycleResearcher closes the loop between
automated research and automated review~\citep{weng2025cycleresearcher}. More
recently, AutoSci proposed a memory-centric system spanning the complete scientific
lifecycle, and FARS reported large-scale deployment of fully automated research
workflows across 67 fine-grained AI/ML topics~\citep{qian2026autosci,tang2026fars}.

A complementary line focuses on scientific ideation and collaboration.
ResearchAgent iteratively generates and reviews ideas grounded in the literature
~\citep{baek2024researchagent}; SciAgents combines knowledge graphs with multi-agent
reasoning for materials discovery~\citep{ghafarollahi2024sciagents}; and Co-Scientist
uses a multi-agent generate--debate--evolve loop for biomedical hypothesis discovery
~\citep{gottweis2025coscientist}. AgentRxiv studies cumulative research through a
shared repository of prior work~\citep{schmidgall2025agentrxiv}, while Arbor uses a
persistent hypothesis tree and isolated executors to refine long-running research
programs~\citep{jin2026arbor}.

\systemname belongs to this autonomous-research family but targets the runtime
layer beneath any single scientific workflow. It maintains a durable campaign,
assigns separate authority to Manager, Planner, Engineer, and Reviewer, and carries
accepted results, failed routes, tools, skills, and task definitions across bounded
missions. The benchmark suite evaluates software engineering, GPU optimization,
model training, research tasks, and data synthesis; a separate mathematical vertical
trace examines proof-oriented research depth.

\subsection{Long-Horizon Agent Runtimes and Memory}

Tool-using agents established the basic interaction loop. Toolformer learns when to
invoke external tools~\citep{schick2023toolformer}, ReAct interleaves reasoning with
environment actions~\citep{yao2023react}, and SWE-agent and OpenHands expose
repository-scale execution through agent--computer interfaces
~\citep{yang2024sweagent,wang2024openhands}. These systems provide the execution
substrate for long tasks; autonomous research additionally requires continuity
across many sessions, experiments, and changes of direction.

Persistent-memory systems address this continuity from several angles. Generative
Agents combine stored observations with reflection and retrieval
~\citep{park2023generative}; Reflexion retains feedback across attempts
~\citep{shinn2023reflexion}; Voyager accumulates executable skills
~\citep{wang2023voyager}; and MemGPT manages multiple memory tiers
~\citep{packer2023memgpt}. Agent Workflow Memory induces reusable procedures from
past trajectories~\citep{wang2024workflowmemory}, while A-MEM organizes memories
through agentic linking and retrieval~\citep{xu2025amem}. \systemname combines
these ideas with explicit ownership: execution produces candidate updates, review
commits the retained form, and the complete event stream remains separate from the
bounded working checkpoint.

\subsection{Learning from Research Trajectories}

Intermediate trajectories are also valuable training material. Process supervision
can train stronger mathematical verifiers than outcome-only supervision
~\citep{lightman2023verify}. AgentInstruct generates post-training corpora through
agentic flows, Agent-FLAN studies data and tuning strategies for agent behavior,
and Agent Lightning converts existing agent trajectories into reinforcement-learning
transitions~\citep{mitra2024agentinstruct,chen2024agentflan,luo2025agentlightning}.
Test-time-compute studies further show that search and verification should be
allocated according to task difficulty rather than by a fixed sampling rule
~\citep{snell2024testtime}.

The trajectories retained by \systemname connect model outputs to concrete states,
tool actions, measurements, artifacts, review decisions, and runtime updates. They
therefore serve two purposes: coordinating the next mission online and forming a
structured corpus for later supervised, preference-based, or reinforcement learning.

\subsection{Evaluation of Research Agents}

Software and general-agent benchmarks established repository-level issue repair
and multi-environment analytical evaluation as core Agent capabilities
~\citep{jimenez2024swebench,liu2023agentbench,ma2024agentboard}.
Research-agent benchmarks increasingly test complete research behavior rather than
isolated question answering. MLE-bench packages Kaggle competitions as
machine-learning engineering tasks~\citep{chan2024mlebench}; RE-Bench compares AI
agents with human experts in open-ended research engineering
~\citep{wijk2024rebench}; AARRI-Bench evaluates granular tasks from the research
lifecycle~\citep{wang2026aarri}; and AIRS-Bench targets frontier research-science
agents across multiple disciplines~\citep{lupidi2026airsbench}. Together they
motivate evaluation in task-native units, with explicit resource accounting and
trajectory-level analysis.

This report combines that benchmark view with a continuous system trace. The
benchmark suite measures outcomes across seven arenas, while the SWE-Bench Pro and
mathematical studies expose runtime evolution, review-driven correction, and
cross-mission research progress.

\section{Problem Formulation}
\label{sec:problem}

\subsection{Dense-Intelligence Tasks}

The public \systemname formulation defines a \emph{dense-intelligence task} as one that
sustains high-frequency reasoning, tool use, verification, and iteration across a
continuous time window until it produces a measurable result~\citep{argus_site}.
Three conditions determine whether a task has this shape:

\begin{enumerate}[label=\textbf{T\arabic*.}]
  \item \textbf{Invention.} The answer is not directly retrievable from a manual or
    database; it must be discovered through search and feedback.
  \item \textbf{Long horizon.} One pass is insufficient. Progress requires many
    dependent iterations over hours or days.
  \item \textbf{Verifiability.} A task-native evaluator can distinguish genuine
    improvement from a persuasive but incorrect claim.
\end{enumerate}

The defining loop is therefore proposal $\rightarrow$ execution $\rightarrow$
measurement $\rightarrow$ revised proposal. Performance engineering, security
research, scientific search, and quantitative research differ in domain but share
this operational structure. The task definition excludes both open-ended ideation
without an evaluator and fixed workflows that require time but no new decisions.

We model a campaign by an objective $o$, an initial artifact $y_0$, an evaluation
procedure $f$, and a resource budget $B$. The artifact may be code, a model, a
dataset, an experimental harness, a proof, or a manuscript. The campaign is divided
into bounded missions indexed by $t$, each producing

\begin{equation}
  \tau_t=\bigl((s_{t,j},a_{t,j},y_{t,j},e_{t,j},r_{t,j})\bigr)_{j=1}^{n_t},
  \label{eq:trajectory}
\end{equation}

where round $j$ has state $s_{t,j}$, actions $a_{t,j}$, resulting artifact state
$y_{t,j}$, observed measurements $e_{t,j}$, and recorded review or admission outcome
$r_{t,j}$. A mission
must terminate, pause, or transfer control explicitly; continuity belongs to the
persistent campaign state rather than an unbounded model session.

\begin{paperbox}{Report-level contract model}
For analysis, we summarize the working task as
$K_t=(\iota,o_t,c_t,v_t)$: standing intent, current objective, known constraints,
and verification criteria. $X_t$ denotes only the clarifications, priorities, and
unresolved questions made explicit at the user interface. A material refinement is
written compactly as
\[
  (X_{t+1},K_{t+1})
  =\operatorname{ManagerAdmit}(X_t,K_t,K'_t,e_t,r_t,u_t),
\]
where $K'_t$ is the proposed contract, $e_t$ is evidence, $r_t$ the recorded
admission outcome, and $u_t$ the authorized operator or Manager
decision. \code{ManagerAdmit} is a normative operator, not one atomic production
API: it projects over current \code{GoalContract} revisions, persisted operator
questions, Planner replanning, Manager Stage/routing transitions, and append-only
provenance.
\end{paperbox}

\begin{figure}[H]
  \centering
  \includegraphics[width=0.95\linewidth]{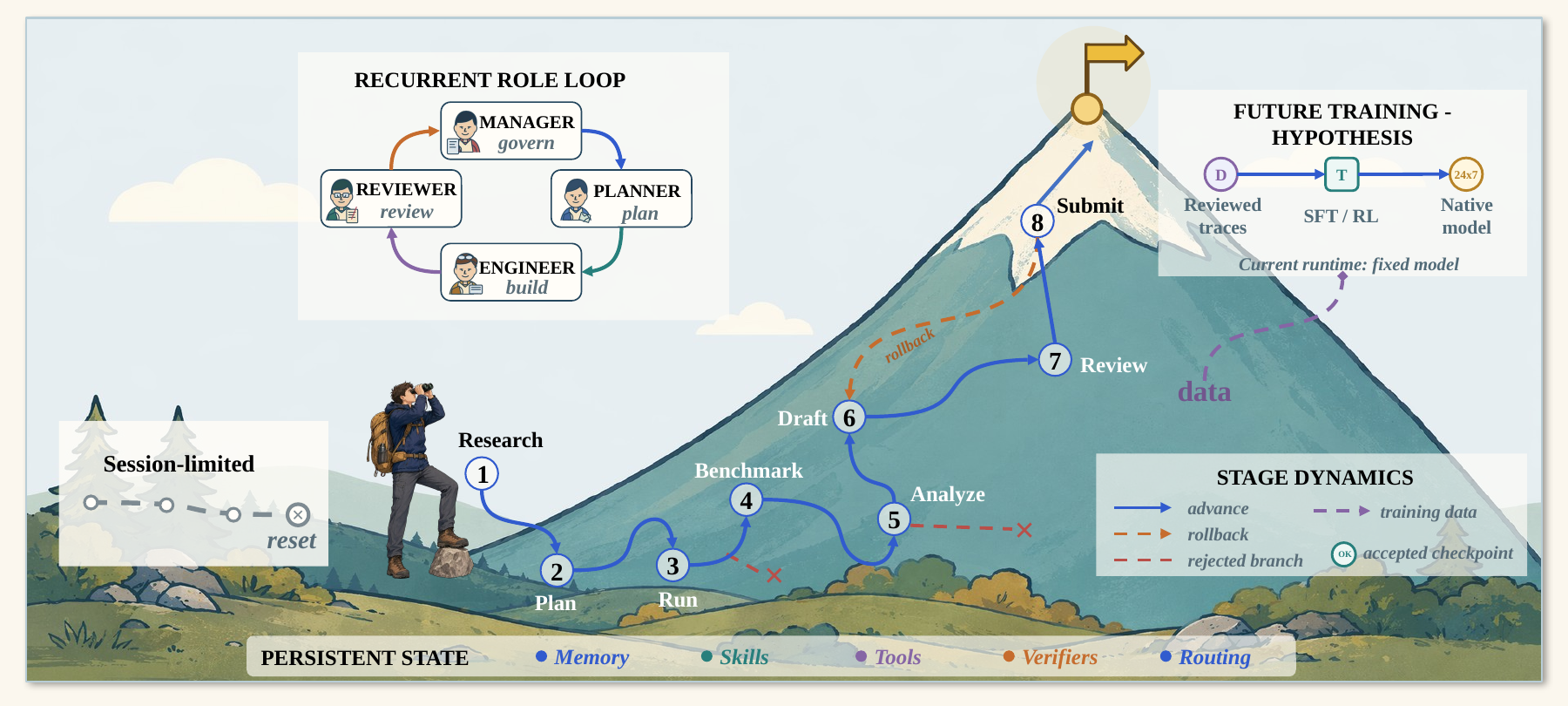}
  \caption{Operational model of long-horizon runtime self-evolution. The left card
  shows a session-limited route ending in reset. The central recurrent role loop is
  reused across eight Manager-controlled research Stages, while the blue route,
  orange rollback, and red rejected branches show that progress need not be
  monotone. Reviewed memory, Skills, tools, verifiers, and routing persist across
  missions. The purple path to future supervised fine-tuning and reinforcement
  learning (SFT/RL) is a research hypothesis; the current
  runtime changes persistent state while keeping model parameters fixed. Geometry
  is conceptual rather than an empirical scale.}
  \label{fig:horizon-mountain}
\end{figure}

Figure~\ref{fig:horizon-mountain} contrasts a session-limited trajectory with the
persistent runtime model. A session-limited agent eventually spends its budget on
reconstruction and repeated work. \systemname instead instantiates a recurrent
Manager--Planner--Engineer--Reviewer loop inside each campaign Stage. Reviewed
state persists below the loop, while Manager-controlled advance, rollback, and
rejected branches change the route taken by later missions.

\subsection{Role-State and Stage Dynamics}

The roles are not four sequential stations that execute once. Within a stage, the
campaign repeatedly applies the transition relation

\begin{equation}
  M \rightarrow P \rightarrow E \rightleftarrows R \rightarrow M,
  \label{eq:role-state-machine}
\end{equation}

where $M$, $P$, $E$, and $R$ denote Manager, Planner, Engineer, and Reviewer.
Reviewer continuation returns work to Engineer; an accepted or blocked verdict
returns control to Manager; and a Manager hold or rollback reopens planning. The
relation is an explanatory state-machine model of the implemented control flow, not
a claim that every transition must invoke a fresh model session.

Let $g_n$ be the current stage after the $n$th Manager decision. Legal transitions
are

\begin{equation}
  g_{n+1}
  \in \{g_n,\operatorname{next}(g_n)\}
  \cup \operatorname{prev}(g_n),
  \label{eq:stage-transition}
\end{equation}

corresponding to hold, advance to the immediate next stage, or rollback to an
earlier stage. For the research vertical, the ordered stages are research, plan,
benchmark, run, analysis, draft, review, and submission. Stage index and task-native
quality need not improve on every cycle: experiments can fail, review can reject a
branch, and later measurements can invalidate earlier assumptions. The intended ascent
is progress in the accepted frontier over many cycles, not monotonic improvement at
each transition.

\subsection{Dense-Intelligence Density}

The website expresses useful intelligence per unit wall-clock time as

\begin{equation}
  \rho_I(T)
  =\frac{1}{T}\int_0^T
  \dot N_{\mathrm{tok}}(t)\,
  \eta_r(t)\,\eta_a(t)\,\eta_v(t)\,dt.
  \label{eq:dense-density}
\end{equation}

$\dot N_{\mathrm{tok}}(t)$ is the instantaneous rate of token-mediated reasoning
and action. The factors $\eta_r(t)$, $\eta_a(t)$, and $\eta_v(t)$ represent the
fractions converted into relevant reasoning, effective action, and valid
verification. The product is intentionally stricter than token throughput: a run
that repeatedly reads the same state, emits activity without changing an artifact,
or accepts unchecked claims can consume many tokens while contributing little to
$\rho_I$.

\paragraph{Operational efficiency factors.}
The three factors describe how work is distributed within an analysis window. For
$W$, let $\mathcal{U}_k(W)$ be the attributable units for
$k\in\{r,a,v\}$: observable reasoning segments, executed actions, or verification
segments. We define
\begin{equation}
  \eta_k(W)
  =\frac{\sum_{u\in\mathcal{U}_k(W)} w(u)q_k(u)}
         {\sum_{u\in\mathcal{U}_k(W)} w(u)},
  \qquad 0\leq q_k(u)\leq1,
  \label{eq:operational-efficiency}
\end{equation}
where $w(u)$ is the attributable Token count for reasoning and verification, and
either one action or its measured cost for action efficiency. The attribution label
$q_k(u)$ records conversion into the intended function:

\begin{itemize}
  \item \textbf{Reasoning efficiency $\eta_r$.} A direct unit derives, revises, or
    rules out a named claim, lemma, blocker, or decision. Repeated context,
    narrative transitions, and notation-only rewriting are transition cost.
  \item \textbf{Action efficiency $\eta_a$.} A direct action changes an artifact,
    result set, accepted decision, or explored branch. A falsification action is
    productive when it certifies a dead branch; retries and interrupted no-op work
    are tracked separately. State synchronization and packaging form the auxiliary
    action category.
  \item \textbf{Verification efficiency $\eta_v$.} A direct verification unit
    compares a named claim with an explicit criterion or artifact, reruns an
    independent check, identifies a concrete defect, or issues a scoped verdict.
    Schema repair, reformatting, and checkpoint canonicalization enable review but
    belong to verification overhead rather than direct checking.
\end{itemize}

We report a direct-work value and an inclusive value that also counts successful
coordination and state maintenance. Equation~\ref{eq:dense-density} uses the three
factors as a multiplicative bottleneck model. Section~\ref{sec:vertical-trace}
measures them in the mathematical campaign, and
Appendix~\ref{app:formula-instantiation} gives the full calculation.

\subsection{Online Life and Runtime Evolution}

The second website object separates model parameters from the operating state
around the model:

\begin{equation}
  \begin{aligned}
    H_{t+1} &= U(H_t,\tau_t,E_t,K_{t+1}), \\
    \theta_{t+1} &= \theta_t.
  \end{aligned}
  \label{eq:runtime-update}
\end{equation}

with the public state definition

\begin{equation}
  H_t=\left\{\text{Memory},\text{Skills},\text{Tools},
  \text{Verifiers},\text{Routing}\right\}.
  \label{eq:website-state}
\end{equation}

$E_t$ is the subset of results admitted from mission $t$. The equality for
$\theta$ is a scope condition: online evolution does not require a gradient update
to the underlying model. Section~\ref{sec:method} refines $H_t$ into named ownership
surfaces, including task/evaluation definitions $Q_t$; subsequent updates treat
$Q_t$ as a component of $H_t$.
The contract argument records the admitted report-level projection. The update $U$
is partial; many missions change only memory, and some change no reusable component.

\subsection{Capability-Set Expansion}

Let $C_t$ denote the effective capability set available at time $t$. The website
writes verification-gated admission and the desired breadth/depth behavior as

\begin{equation}
  \begin{aligned}
    C_{t+1}
      &= C_t\cup\{c:\operatorname{Verify}(c,E_t)\geq\epsilon\}, \\
    W_{t+1} &\geq W_t,
    & D_{t+1}(d) &\geq D_t(d).
  \end{aligned}
  \label{eq:ood-expansion}
\end{equation}

where $W_t$ is frontier width across domains and $D_t(d)$ is depth within domain
$d$. The inequalities define retention desiderata over validated capability.
Skills can become stale, verifiers can be revised, and performance can regress on
harder instances. A negative result can still add value by excluding a failed branch even when the
effective capability set does not grow.

\subsection{Research Value}

Dense activity matters only when it produces new, valid, reusable information. The
website therefore defines the accumulated research value over horizon $T$ as

\begin{equation}
  V_R(T)
  =\int_0^T
  \rho_I(t)\,\Delta I(t)\,p_{\mathrm{valid}}(t)\,
  \eta_{\mathrm{reuse}}(t)\,dt,
  \label{eq:research-value}
\end{equation}

where $\Delta I(t)$ is information gain, $p_{\mathrm{valid}}(t)$ is the probability
that the claimed increment survives review, and
$\eta_{\mathrm{reuse}}(t)$ is the fraction that remains useful for later work.
Equation~\ref{eq:research-value} separates Token volume from the properties that
make a result useful: information gain, validity, and reuse. The experiments measure
these components in their native task units.

\subsection{The Information Advantage of Process Data}

Finally, the public formalization distinguishes the accepted artifact from the
trajectory that produced it:

\begin{equation}
  D_{\mathrm{process}}
  =\{(s_k,a_k,e_k,r_k,\Delta H_k)\}_{k=1}^{N}
  \supsetneq
  D_{\mathrm{final}}=\{y^\star\}.
  \label{eq:process-data}
\end{equation}

The process record contains states, actions, measurements, review feedback, and
runtime-state changes, including failed branches omitted from $y^\star$. This
additional information can reduce repeated search, guide later analysis, and provide
grounded material for later skill construction or evaluation. \systemname retains
typed, privacy-preserving observations and verdicts rather than private
chain-of-thought transcripts.

The set inclusion in Equation~\ref{eq:process-data} can be strengthened into a
decision-theoretic statement. Let $P$ denote a typed process record and let
$Y=g(P)$ be its final-artifact projection. For a downstream task $q$ with target
$Z_q$, loss $\ell_q$, and decision policy $\pi$, define the minimum achievable risk

\begin{equation}
  \mathcal{R}_q(X)
  =\inf_{\pi}\mathbb{E}\!\left[\ell_q\!\left(\pi(X),Z_q\right)\right].
  \label{eq:downstream-risk}
\end{equation}

\begin{paperbox}{Proposition 1: process-data dominance}
If $Y=g(P)$, then $\mathcal{R}_q(P)\leq\mathcal{R}_q(Y)$ for every downstream
decision problem $q$. The inequality is strict for some $q$ whenever two process
records can yield the same final artifact while implying different optimal next
actions. The proof is immediate: every policy using $Y$ can be reproduced from
$P$ by first applying $g$, while the reverse simulation need not exist. This is the
Blackwell-information ordering specialized to a research trajectory
~\citep{blackwell1953equivalent}.
\end{paperbox}

Dominance is informational, not computational. A raw trajectory can be too large,
stale, or contradictory to use efficiently. For a context budget $b$, the relevant
object is therefore a typed compression $\psi(P)$ that trades downstream risk
against retrieval and validation cost:

\begin{equation}
  \psi_b^\star
  =\arg\min_{\psi:\,\operatorname{size}(\psi(P))\leq b}
  \mathbb{E}_{q\sim\mathcal{D}}
  \left[\mathcal{R}_q(\psi(P))+\lambda_c C_{\mathrm{read}}(\psi(P))\right].
  \label{eq:process-compression}
\end{equation}

Equation~\ref{eq:process-compression} explains why an append-only event tape and a
bounded reviewed checkpoint serve different purposes. The tape preserves the more
informative experiment; the checkpoint approximates a decision-useful compression
under a finite context budget. Failed branches belong in the compression when they
change the next optimal action, not merely because they occurred.

\subsection{Reusable State and Compounding Intelligence}

Process data compounds only when an admitted state update improves later work. Let
$H_t\oplus\Delta H_t$ denote the state after accepting mission $t$, and let
$q_{t+1:t+L}$ be the next $L$ tasks. We define the discounted reuse value

\begin{equation}
  G_L(\Delta H_t)
  =\sum_{j=1}^{L}\gamma^{j-1}
  \left[
    \mathcal{R}_{q_{t+j}}(H_t)
    -\mathcal{R}_{q_{t+j}}(H_t\oplus\Delta H_t)
  \right].
  \label{eq:reuse-gain}
\end{equation}

$G_L>0$ means that the accepted memory, skill, verifier, routing rule, or certified
dead branch reduces future loss under a prespecified task distribution. $G_L<0$
captures negative transfer. This definition turns ``compounding intelligence''
from monotone rhetoric into a counterfactual claim: reuse must outperform a frozen
state on matched future tasks. Workflow-memory experiments show that induced
reusable routines can improve success and reduce steps
~\citep{wang2024workflowmemory}. Section~\ref{sec:discussion} connects this quantity
to the current \systemname traces.

The website's density and research-value equations can then be joined into a
protocol-specific \emph{verified reusable yield}:

\begin{equation}
  \mathcal{Y}_{\mathrm{PRI}}(W)
  =\frac{
    \sum_{t\in W}p_{\mathrm{valid},t}
    \left[\Delta I_t+\lambda_g G_L(\Delta H_t)\right]
  }{
    \sum_{t\in W}N_{\mathrm{tok},t}
  }.
  \label{eq:verified-reusable-yield}
\end{equation}

$\Delta I_t$ is immediate task-native information gain, such as a score
improvement, theorem strengthening, or certified branch elimination;
$p_{\mathrm{valid},t}$ is estimated by an external evaluator or a calibrated review
procedure; and $G_L$ measures future reuse. $\lambda_g$ converts immediate and
future value into one protocol-specific unit. Equation~\ref{eq:verified-reusable-yield}
connects immediate research output to future reuse. The current substitutions are
reported in Appendix~\ref{app:formula-instantiation}.

\subsection{Review as Selective Error Correction}

Let $C$ denote whether a proposed state increment is correct, $A$ whether the
Reviewer accepts it, $p=\Pr(C=1)$ the proposal base rate,
$\alpha=\Pr(A=1\mid C=1)$ Reviewer sensitivity, and
$\beta=\Pr(A=1\mid C=0)$ false-acceptance rate. Bayes' rule gives

\begin{equation}
  \Pr(C=1\mid A=1)
  =\frac{\alpha p}{\alpha p+\beta(1-p)}.
  \label{eq:review-precision}
\end{equation}

If $\alpha>\beta$, accepted state is more precise than the proposal stream; review
acts as a selective error-correction channel. Its operating point balances accepted
precision, recall, and Token cost. Section~\ref{sec:results} reports the observed
repair and recovery rates.

\subsection{Role-Resolved Vertical Traces}

Endpoint metrics do not show how an Agent system actually conducts research. We
therefore project each mission trajectory in Equation~\ref{eq:trajectory} onto
the four role-owned actions

\begin{equation}
  \phi(\tau_t)=\left(m_t,p_t,x_t,r_t,\Delta H_t\right),
  \label{eq:role-resolved-trace}
\end{equation}

where $m_t$ is the Manager's objective or stage decision, $p_t$ is the Planner's
bounded task and dependency decision, $x_t$ is the Engineer's artifact-producing
execution, $r_t$ is the structured completion decision together with its source,
and $\Delta H_t$ is the durable state update exposed to later missions. The
decomposition makes two authority constraints explicit: only the Manager changes
campaign stage, while mission completion records either an allowed Engineer
self-review or a required/requested independent Reviewer verdict. Vertical policy
and stage-closing tasks prevent Engineer self-certification where independent
review is mandatory; the Planner may propose future work but cannot certify the
mission.

A vertical trace is useful when these tuples remain linked across many missions.
It shows whether the objective survived failures, whether the Planner consumed
previously accepted state, whether the Engineer changed real artifacts, whether
the Reviewer rejected or redirected incomplete work, and which decisions became
inputs to the next mission. This explanatory model organizes the Agent workflow;
Section~\ref{sec:vertical-trace} applies it to one mathematical campaign.

\section{\systemname Runtime}
\label{sec:method}

\subsection{System Overview}

\systemname is a general-purpose agentic runtime built around four model-driven roles
and three system planes. The \emph{control plane} anchors the campaign and schedules
work; the \emph{execution plane} performs one bounded mission against real tools
and artifacts; and the \emph{record plane} stores what happened without deciding
whether the work is scientifically complete. The distinction is operational:
control owns scheduling, execution owns work and mission review, and the record plane owns
the immutable record but no completion decision.

The four roles divide research authority as summarized in
Table~\ref{tab:roles}. The Manager spans the campaign rather than one execution
round. The Planner authors future work and the Engineer performs it. Mission-level
completion has an explicit source: allowed low-risk bounded work may use Engineer
self-review, while vertical policy, stage-closing tasks, or the Engineer's own
request require an independent Reviewer. The interfaces are structured objects
rather than untyped prose, which makes ownership and recovery behavior explicit.

\begin{table}[H]
\centering
\small
\begin{tabularx}{\linewidth}{@{}p{1.6cm} p{4.0cm} X@{}}
\toprule
\rowcolor{papersand}
\textbf{Role} & \textbf{Primary responsibility} & \textbf{Authoritative output} \\
\midrule
Manager & Commit the objective, lifetime, route, and campaign stage & Campaign division and stage transition \\
Planner & Decompose the current research state into bounded tasks and dependencies & Task specifications, plan status, and stage checklist updates \\
Engineer & Modify artifacts, invoke tools, and run experiments for one round & Execution result, artifacts, measurements, handoff, and structured \code{review=skip|required} selection \\
Reviewer & Independently inspect artifacts and execution records when required or requested & \code{done}, \code{continue}, or \code{blocked}, plus grounded next action and retained-state edits \\
\bottomrule
\end{tabularx}
\caption{Role separation in \systemname. The Engineer executes and may self-review
allowed low-risk bounded work; mandatory or requested independent review belongs to
the Reviewer.}
\label{tab:roles}
\end{table}

\begin{paperbox}{Algorithm 1: Campaign scheduling and the reviewed mission loop}
\small
\begin{tabbing}
\quad\=\quad\=\quad\=\quad\=\kill
\textbf{Input:} standing intent $\iota$, contract $K_t$, runtime state $H_t$, backlog $B$, budget $b$ \\
$c \leftarrow \operatorname{ManagerCommit}(\iota,K_t)$; persist campaign identity \\
\textbf{while} $b$ remains and $c$ is not complete \textbf{do} \\
\> \textit{// campaign level: the mission boundary is fixed here} \\
\> $N \leftarrow \operatorname{StableTopological}(\operatorname{Planner}(B,H_t,c,K_t))$ \\
\> \textbf{for} each bounded mission $q_0 \in N$, enqueued one at a time \textbf{do} \\
\>\> $L \leftarrow \operatorname{ExposeSkillLibraries}(q_0)$; \textit{paths only, retrieval is the agent's} \\
\>\> $\Gamma \leftarrow [\,]$; $q \leftarrow q_0$ \\
\>\> \textbf{repeat} \textit{// mission level: $q_0$ is now fixed for every round below} \\
\>\>\> $(y,e,d) \leftarrow \operatorname{Engineer}(q,H_t,L)$ \\
\>\>\> \textbf{if} $d=\code{skip}$ and $\operatorname{SelfReviewAllowed}(q)$ \\
\>\>\>\> $r \leftarrow \operatorname{EngineerSelfReview}(q,y,e)$ \\
\>\>\> \textbf{else} $r \leftarrow \operatorname{Reviewer}(q,y,e)$ \textit{; scoped to this round} \\
\>\>\> append $(q,y,e,r,r.\code{source})$ to $\Gamma$ \\
\>\>\> \textbf{if} $r=\code{continue}$ \textbf{then} $q \leftarrow \operatorname{AdaptAfterRejections}(\Gamma)$ \\
\>\> \textbf{until} $r\in\{\code{done},\code{blocked},\code{paused}\}$ or a governance threshold fires \\
\>\> $(\code{status},\code{reason}) \leftarrow \operatorname{PreSettlementGuard}(\Gamma,\code{status},\code{reason})$ \\
\>\> $\tau \leftarrow \operatorname{TraceProjection}(\Gamma)$; \quad $E \leftarrow \operatorname{AdmitResult}(\tau)$ \\
\>\> $H_{t+1} \leftarrow \operatorname{SettleMissionOutcome}(H_t,\tau,E)$ \textit{; learning happens here} \\
\> \textbf{if} evidence proposes a contract change $K'_t$ \textbf{then} \\
\>\> $(K_{t+1},\rho_t) \leftarrow \operatorname{ReviseContract}(K_t,K'_t,\code{by},\code{confirmation})$ \\
\> $t \leftarrow t+1$; refill $B$ when required \\
\textbf{end while} \\
\textbf{Output:} accepted artifacts, inspectable trajectory, admitted contract, and updated runtime state
\end{tabbing}
\end{paperbox}

The algorithm above separates two levels that are easy to conflate. The
Planner runs at the campaign level and fixes the mission boundary $q_0$; the round
loop beneath it varies $q$ only through \code{AdaptAfterRejections}, which reacts to
Reviewer rejections \emph{within} that boundary. No component inside the round loop
can rewrite $q_0$. Section~\ref{sec:endogenous-harness} treats the consequence of
that asymmetry as a first-class failure mode rather than an implementation detail.

Three further details matter because they differ from the idealized loop.
\emph{Skill libraries are exposed, not injected}: the runtime publishes library paths
and the acting agent performs its own retrieval, so no matcher decides in advance
which prior experience is relevant. \emph{Termination is governed by named
thresholds} rather than by Reviewer verdicts alone: a maximum round count, a
no-progress threshold, a soft round limit, a hard escalation count, and a backend
failure threshold each end or escalate a mission, and a pre-settlement guard may
override the recorded status before anything is learned. \emph{Learning is settled
per mission, not per round}: reusable state changes once, at the mission outcome.

\code{ReviseContract} is where evidence becomes an authorized change of target, and
its gate is deliberately two-tier. A \code{GoalContract} holds two kinds of clause.
Semantic clauses, exclusions, and recorded ambiguities move freely, because
clarifying what the user meant is ordinary Manager work. Precise clauses and the
objective itself require a confirmation covering exactly the clause identifiers that
change; without it the revision is refused. That split is the implemented form of the
distinction Section~\ref{sec:problem} draws analytically: intent is stable, wording
is negotiable, and the operational target moves only with recorded authority.

\subsection{Bounded Missions over Durable State}

The long-lived object in \systemname is the campaign, not a provider transcript. A
campaign has a persisted identity and objective; its work is divided into missions
that have explicit outcomes. The scheduler assigns one mission at a time, records
its result, and advances only at a clean mission boundary. Mission assignment is
transactional, preventing duplicate work under concurrency, and resumed execution
is tied to the persistent campaign identity.

Engineer and Reviewer calls use fresh provider sessions for each round. Cross-session
continuity comes from an ordinary shared \code{CHECKPOINT.md} containing durable
state, evidence references, open questions, and the next step. This follows the
broader move from monolithic context toward managed memory tiers and reusable
workflows~\citep{packer2023memgpt,wang2024workflowmemory,xu2025amem}. The Engineer
updates the checkpoint after execution. When independent review is invoked, the
Reviewer reads and corrects the same file and is its final editor for that round;
on an accepted self-review path, the Engineer's version remains the handoff. The
full history remains in artifacts and the event record.

This design makes restart and upgrade behavior part of the method. A running process
hands off only at a mission boundary, and the replacement resumes from the same
persistent campaign identity. After an interruption, replay or reassignment begins
from committed campaign state rather than reconstructing the task from a model
transcript.

\subsection{Review and Completion}

All role invocations pass through a common instrumentation layer that records usage
and associates each call with the mission trace. The Engineer and Reviewer share
the same artifact state, allowing review of the actual outputs and execution record
rather than only the Engineer's summary. A typed, append-only trace is the canonical
timeline; user interfaces are projections over that record.

The source of a mission completion verdict is explicit. When self-review is enabled
and no vertical or task policy requires independence, an Engineer may select
\code{review=skip}; its prompt restricts that choice to low-risk bounded work with
a passing verifier. The runtime accepts the explicit agent judgment without adding
a second heuristic or validator and records
\code{review\_source=engineer\_self\_review}. Otherwise a fresh Reviewer
returns the structured verdict, and stage-closing or vertical-required review cannot
be waived. Completion never comes from filenames, Token volume, or prose keywords.
The corresponding artifacts and execution records remain inspectable, while
credential redaction and result provenance preserve the selected verdict source.

\subsection{Verification-Gated Fixed-Model Runtime Self-Evolution}

Equation~\ref{eq:runtime-update} separates the model from the state around it.
Table~\ref{tab:runtime-ownership} identifies how each component can change and who
owns the committed update. The ownership model is intentionally not uniform. Memory
and skills use a work-versus-certification split: the Engineer or Scientist produces
a candidate, and the Reviewer commits the retained form. Tools and procedures are
system-configured. Stage checklists are Planner-owned, with Reviewer feedback rather
than a second commit gate. Routing policy is Manager-committed, while task
definitions are Planner-authored and scheduler-committed.

We use \emph{verification-guided} for the overall control policy that decides whether
to persist, stop, or pivot, and \emph{verification-gated} for the narrower admission
condition on reusable updates. We use \emph{runtime self-evolution} in a deliberately
narrow, fixed-model sense.
The underlying model parameters remain unchanged; what evolves is the persistent
state that later missions retrieve or obey. \emph{Verification-gated} is the
umbrella term used in this report: a
generated candidate is not reusable merely because a role produced it. Admission
requires the task-native evidence that exists for that surface and a commit by its
authorized owner. Depending on policy and risk, this can combine an official
executable verifier with independent Reviewer judgment, permitted Engineer
self-review, or a Manager, Planner, Scheduler, or system-configuration commit. It
does not mean that every low-risk update requires an independent Reviewer. We use
\emph{Reviewer-gated} only for the independent-Reviewer path and \emph{external
grader} only for a task-native evaluator outside the role loop.

A complete update cycle has four parts:
(1) an execution trajectory produces a candidate memory, skill, procedure,
verification rule, routing decision, or task definition; (2) the responsible role
checks the candidate against artifacts and task-native evidence; (3) the authorized
owner commits, revises, or rejects the update; and (4) a later mission retrieves the
retained state as part of its starting context or execution policy. Activity that
does not survive this commit-and-reuse path is not counted as self-evolution.

\begin{table}[H]
\centering
\small
\begin{tabularx}{\linewidth}{@{}p{2.25cm} p{2.75cm} p{2.55cm} X@{}}
\toprule
\rowcolor{papersand}
\textbf{State} & \textbf{Change source} & \textbf{Commit owner} & \textbf{Persistent form} \\
\midrule
$M$: memory & Engineer trajectory & Reviewer & Curated checkpoint and event-derived journal \\
$S$: skills & Engineer / Scientist & Reviewer & Versioned skill library \\
$A$: tools and procedures & System configuration & System configuration & Versioned tool and procedure registry \\
$V$: verification & Planner & Planner & Stage checklist; Reviewer supplies feedback \\
$R$: routing and roles & Runtime policy & Manager & Campaign routing policy \\
$Q$: tasks and evaluations & Planner & Scheduler & Backlog task specifications and scopes \\
\bottomrule
\end{tabularx}
\caption{Attributable ownership of fixed-model runtime evolution. A mission usually
updates only a subset of these components.}
\label{tab:runtime-ownership}
\end{table}

Two knowledge surfaces carry reusable experience. Versioned skills store procedures
that can be matched to later tasks. A project knowledge base stores source-linked
records and synthesized pages derived from reviewed outcomes. Neither surface is
treated as automatically correct: entries can be revised, archived, or retired when
later results contradict them.

This mechanism can improve a later mission without changing the model itself. A
verified retained failure can prevent a repeated dead end; an admitted procedure can shorten
environment inspection; a verifier can reject a previously accepted shortcut; and
a routing update can assign independent review to a higher-risk task. The mechanism
does not imply monotonic improvement. Some missions commit no reusable state,
retained state can become stale, and a harder task distribution can increase cost
even after useful state has accumulated.

\subsection{Reliability and Resource Governance}

Long-running systems fail through operational drift even when individual model
calls are strong. \systemname therefore combines round-level progress
classification with mission-boundary replanning. Work that repeatedly produces no
decision or artifact change is stopped or reformulated, while long-running external
jobs are tracked separately from model reasoning. These mechanisms bound execution;
scientific correctness remains with the task evaluator and the explicitly selected
Engineer-self-review or independent-Reviewer judgment path.

Resource accounting is centralized at model-call and external-job boundaries.
Budgets are checked before work begins and reconciled after completion, preventing
individual roles from expanding their own allocation. Concrete scheduling,
sandboxing, and deployment mechanisms are implementation choices rather than part
of the scientific claim.

\section{Empirical Methodology}
\label{sec:methodology}

We evaluate \systemname across seven benchmark arenas spanning software repair,
GPU-kernel optimization, language-model training, training-speed optimization,
research-assistant tasks, and data synthesis. Each benchmark is reported in its
native unit. SWE-Bench Pro is one benchmark in this suite; its sequential task log
also enables deeper analyses of runtime evolution and Reviewer intervention.

\subsection{Research Questions}

\paragraph{RQ1: Benchmark performance.}
What task-native outcomes does \systemname achieve across the seven benchmark
arenas, and how do they compare with the reference reported for each arena?

\paragraph{RQ2: Fixed-model runtime self-evolution.}
Within the SWE-Bench Pro run, how does solve-time resource demand change as reviewed
Skill, Wiki, verification, and routing state accumulates while the underlying model
parameters remain fixed?

\paragraph{RQ3: Reviewer routing and recovery.}
Within the same SWE-Bench Pro run, how often is an independent Reviewer invoked,
and how many initially rejected solutions are recovered after Reviewer feedback?

\paragraph{RQ4: Process-to-capability mechanisms.}
How do the retained traces quantify process-data advantage, verification-gated
correction, Token conversion, and cross-task reuse?

\subsection{Benchmark Suite}

SWE-Bench Pro evaluates repository-level software-engineering tasks with executable
acceptance tests~\citep{deng2025swebenchpro}. The remaining arenas evaluate B200
kernel optimization, nanochat training on B200 and H100, the nanoGPT speedrun,
AARRI-Bench, and Math-Reasoning Data Synthesis from the Arbor suite
~\citep{lin2026solexecbench,karpathy_nanochat,karpathy_autoresearch,
jordan_nanogptbench,wang2026aarri,jin2026arbor}. Because rank, bits per byte,
elapsed time, solve rate, and pass-gap are incompatible metrics, we do not average
them into one score.

\begin{table}[H]
\centering
\footnotesize
\renewcommand{\arraystretch}{1.16}
\begin{tabularx}{\linewidth}{@{}p{2.15cm} X p{3.25cm} p{1.05cm}@{}}
\toprule
\rowcolor{papersand}
\textbf{Benchmark} & \textbf{What the agent must do} & \textbf{Evaluation metric} & \textbf{Better} \\
\midrule
SWE-Bench Pro & Repair a real repository issue by editing production code; the submitted patch must satisfy task-specific executable acceptance tests. & Fraction of tasks resolved by the official verifier & Higher \\
SOL-ExecBench & Optimize real GPU kernels without changing their outputs, targeting execution close to an analytically derived hardware speed-of-light bound. & SOL score, global rank, and high placements after correctness checks & Higher \\
nanochat B200 & Improve the nanochat training program under a fixed five-minute run on one B200 GPU. & Validation bits per byte (BPB), a compression-oriented language-model metric & Lower \\
nanochat H100 & Solve the same five-minute nanochat training objective on one H100; it is a separate hardware-specific optimization problem. & Validation BPB under the H100 protocol & Lower \\
nanoGPT speedrun & Modify the training system so an eight-H100 run reaches validation loss $3.28$ as quickly as possible. & Mean time to the target over ten verifier runs & Lower \\
AARRI-Bench & Complete 82 granular tasks representing the judgment, diligence, and professional practice expected from a research intern. & Number and percentage of tasks solved & Higher \\
Math-Reasoning Data & Build a pipeline that generates AIME-style reasoning problems that are difficult on the first attempt but solvable with additional attempts. & Mean $\mathrm{pass@4}-\mathrm{pass@1}$ gap & Higher \\
\bottomrule
\end{tabularx}
\caption{Definitions of the seven benchmark arenas. B200 and H100 nanochat runs
are kept separate because hardware changes the optimization problem.}
\label{tab:benchmark-definitions}
\end{table}

The six non-SWE arenas use GPT-5.5 through Codex. The SWE-Bench Pro evaluation uses
GPT-5.5/xhigh through Copilot for both Direct Copilot and \systemname. The full
731-task comparison reports approximate accuracy and the aggregate Token ratio
\begin{equation}
  R_{\mathrm{tok}}
  = \frac{N_{\mathrm{\systemname}}^{\mathrm{total}}}
         {N_{\mathrm{Copilot}}^{\mathrm{total}}}
  \approx 1.41.
  \label{eq:aggregate-token-ratio}
\end{equation}
Raw Copilot Token totals and per-Wave resource traces were not retained; therefore
the longitudinal analysis is \systemname-only.

\subsection{SWE-Bench Pro Runtime Self-Evolution Analysis}

For a completed \systemname Wave $w$ with task set $\mathcal{T}_w$, we measure
\begin{align}
  \bar I_w^{\mathrm{solve}}
  &= \frac{1}{|\mathcal{T}_w|}
     \sum_{i\in\mathcal{T}_w} I_i^{\mathrm{solve}},
  \label{eq:solve-token-wave}\\
  \bar T_w^{\mathrm{agent}}
  &= \frac{1}{|\mathcal{T}_w|}
     \sum_{i\in\mathcal{T}_w} T_i^{\mathrm{agent}}.
  \label{eq:solve-time-wave}
\end{align}
$I_i^{\mathrm{solve}}$ contains all model calls in the accepted task trajectory,
including routing, Skill adaptation, execution, and any independent review.
$T_i^{\mathrm{agent}}$ measures active
workflow time. Orchestration wait, environment preparation, external verification,
infrastructure recovery, and post-task knowledge maintenance are excluded.

Completed Waves are summarized by task-weighted windows: W1--6 (startup), W7--12
(early reuse), W13--18 (composition shift), W19--22 (mature operation), and
W23--24 (late difficult tasks). Two incomplete Waves are omitted from the grouped
means. The primary comparison is startup versus mature operation; composition and
difficulty variation remain visible.

The unit of analysis is the observed sequential window, not a matched task pair.
The startup window begins with less project-specific state, whereas later windows
can retrieve reviewed repository knowledge, reusable procedures, prior failure
records, verification guidance, and updated routing decisions. Lower Token or time
demand in a later window is therefore consistent with runtime self-evolution when
the retained state removes repeated inspection or failed work. It is not, by
itself, a causal estimate of the value of that state. Task identity, repository
mix, execution latency, and difficulty also change across Waves, and no matched
frozen-state replay is available.

\subsection{SWE-Bench Pro Reviewer Analysis}

A task is \emph{Reviewer-invoked} when its final trajectory contains an independent
Reviewer model call; otherwise it follows Engineer self-review. A Reviewer verdict
of \code{continue} is a revision request. We report both subsequent official
verifier success and the stricter case in which a later Reviewer verdict is
\code{done}. The latter is called a \emph{strict Reviewer rescue}.

Reviewer routing is adaptive rather than randomized. The analysis reports routing,
revision, and recovery behavior over the observed task sequence.

The 41-artifact research portfolio is reported separately as a measure of research
program breadth.

\subsection{Representative Vertical-Trace Protocol}

To complement endpoint and software-trajectory measurements, we report one
Erd\H{o}s--Gy\'arf\'as mathematical campaign as a representative vertical trace.
The trace is role-led rather than theorem-led: it records what the Manager,
Planner, Engineer, and Reviewer did during recovery, problem selection, route
testing, proof production, revision, acceptance, and state retention. The reported
record contains one accepted route falsification and six proof-backed research
deltas; detailed mathematical statements remain in the supplementary claim table.

The protocol tracks whether the runtime retains a falsified branch, carries named
blockers across missions, and admits later work after artifact-based review. The
role trace, claim table, and figure provenance are released with the
supplementary materials.

\subsection{Theory-to-Measurement Mapping}

We map the theory to four reported quantities. Immediate information gain
$\Delta I_t$ is represented only in task-native units: benchmark improvement,
accepted theorem delta, or certified branch elimination. Review correction is
represented by Reviewer \code{continue}--revise trajectories and later
external-verifier outcomes. Reuse value $G_L$ is approximated observationally by
later Token/time demand as Skill and Wiki state accumulates. Token conversion is
represented by the role- and mission-level attribution in the mathematical case.
Appendix~\ref{app:formula-instantiation} records the corresponding substitutions in
their native units.

\section{Results}
\label{sec:results}

\subsection{Capability Floor Across Seven Arenas}

Table~\ref{tab:benchmark-results} is the primary empirical summary and reports all
seven benchmarks in one comparison; Table~\ref{tab:benchmark-definitions} defines
the task and metric for each row.

\begin{table}[H]
\centering
\scriptsize
\renewcommand{\arraystretch}{1.14}
\begin{tabularx}{\linewidth}{@{}p{1.55cm}p{1.55cm}p{1.10cm}X p{1.75cm}X@{}}
\toprule
\rowcolor{papersand}
\textbf{Benchmark} & \textbf{Backbone} & \textbf{Backend} & \textbf{Protocol} & \textbf{\systemname result} & \textbf{Comparison} \\
\midrule
SWE-Bench Pro & GPT-5.5/xhigh & Copilot & 731 software-repair tasks & $\approx$78\% accuracy & Direct Copilot $\approx$59\%; 1.41$\times$ aggregate Tokens \\
SOL-ExecBench & GPT-5.5 & Codex & B200; 101 kernels & Global \#6; 2$\times$\#1; 7 top-3 & Two head-to-head wins over Recursive \\
nanochat B200 & GPT-5.5 & Codex & 5 min; 1$\times$B200 & 0.9636 BPB & Human best 0.9646 \\
nanochat H100 & GPT-5.5 & Codex & 5 min; 1$\times$H100 & 0.9855 BPB & Human best 0.9879 \\
nanoGPT speedrun & GPT-5.5 & Codex & 8$\times$H100; $N{=}10$ & 79.77 s & Same-device human 80.18 s \\
AARRI-Bench & GPT-5.5 & Codex & 82 research-intern tasks & 63/82 (76.8\%) & Paper best 68.3\% \\
Math-Reasoning Data & GPT-5.5 & Codex & AIME-style synthesis & 28.0 gap & Arbor 20.83; Claude 8.33; Codex 6.25 \\
\bottomrule
\end{tabularx}
\caption{Results across seven benchmark arenas. Values remain in task-native units
and are not cross-normalized. Backbone denotes the model driving the research
agent; Backend denotes the execution surface.}
\label{tab:benchmark-results}
\end{table}

The runtime is not bound to one backbone or one execution surface. A separate
SWE-Bench Pro run replaces GPT-5.5 on Copilot with GLM-5.2 at a 200K-token context
window, driven through Claude Code, and is currently at 70.94\% accuracy. That run
is still in progress and has no matched Direct baseline on this backend, so we do
not enter it in Table~\ref{tab:benchmark-results} or in any longitudinal analysis;
we note it because the runtime carried the same contract, routing, and Reviewer
policy onto a different model and a different agent surface without
re-instrumentation.

As a small upstream-adoption example, an \systemname-optimized TileLang
\code{RWKV6} kernel was reviewed by a Moonshot AI-affiliated FLA collaborator and
merged into \code{fla-org:main}
(\href{https://github.com/fla-org/flash-linear-attention/pull/1045}{pull request (PR) \#1045};
\href{https://github.com/fla-org/flash-linear-attention/commit/c70f11c5530142450525549cc96d13d9f5165f69}
{commit \code{c70f11c}}); Appendix~\ref{app:upstream-kernel} records the technical
details.

An eighth arena is in progress and is reported here as a partial result. The
MLE-Bench Lite campaign runs the official MLE-Bench Low split
~\citep{chan2024mlebench} under a reviewer-approved medal gate: a competition counts
as complete only when an independently reviewed submission earns a Kaggle medal, so
the reported outcomes are medals rather than raw leaderboard positions.
Table~\ref{tab:mle-medals} lists the competitions closed to date: nine medals,
evenly split into three gold, three silver, and three bronze. The gate is narrow
rather than nominal. On \code{denoising-dirty-documents} the campaign settled
0.00009 RMSE short of the silver band, and on
\code{jigsaw-toxic-comment-classification-challenge} it cleared the bronze
threshold by 0.00018 AUC; both outcomes were awarded by the external grader
against the official Kaggle leaderboard, not by internal review. Two further
competitions were graded without reaching a medal: \code{dog-breed-identification}
finished above the median after twelve submissions, and
\code{new-york-city-taxi-fare-prediction} remained below it after four. The
transparent-conductor result is also a verification-gated route change: the
external grader rejected the initial approach at 0.208 RMSLE, the campaign
replaced it with a public-state-of-the-art method, and the grader then certified
the replacement inside the medal band at 0.06402 against a 0.06582 bronze
threshold. The remaining competitions continue to run; we exclude this arena from
Table~\ref{tab:benchmark-results} and will report the full split on completion.

\begin{table}[H]
\centering
\small
\renewcommand{\arraystretch}{1.18}
\begin{tabularx}{\linewidth}{@{}Xp{1.85cm}p{1.75cm}p{1.35cm}@{}}
\toprule
\rowcolor{papersand}
\textbf{Competition} & \textbf{Metric} & \textbf{\systemname} & \textbf{Medal} \\
\midrule
\code{aerial-cactus-identification} & AUC $\uparrow$ & \textbf{1.00000} & \textbf{Gold} \\
\code{dogs-vs-cats-redux-kernels-edition} & LogLoss $\downarrow$ & \textbf{0.02168} & \textbf{Gold} \\
\code{detecting-insults-in-social-commentary} & AUC $\uparrow$ & \textbf{0.91386} & \textbf{Gold} \\
\code{histopathologic-cancer-detection} & AUC $\uparrow$ & 0.98312 & Silver \\
\code{leaf-classification} & LogLoss $\downarrow$ & 0.00196 & Silver \\
\code{mlsp-2013-birds} & AUC $\uparrow$ & 0.91479 & Silver \\
\code{denoising-dirty-documents} & RMSE $\downarrow$ & 0.02618 & Bronze \\
\code{jigsaw-toxic-comment-classification-challenge} & AUC $\uparrow$ & 0.98657 & Bronze \\
\code{nomad2018-predict-transparent-conductors} & RMSLE $\downarrow$ & 0.06402 & Bronze \\
\bottomrule
\end{tabularx}
\caption{MLE-Bench Lite competitions closed with a medal by \systemname{} so far.
Arrows give the direction of improvement: area under the ROC curve (AUC) is
maximized, while log loss (LogLoss), root mean squared error (RMSE), and
column-wise root mean squared logarithmic error (RMSLE) are minimized. Each score
is the Reviewer-approved submission that the external MLE-Bench grader scored, and
each medal is the grader's award against that competition's official Kaggle
leaderboard. Seven of the nine rows are reconciled against the 2026-07-29
reviewer-approved campaign snapshot, which records 40 graded submissions across
nine competitions. The campaign is still running, so this table reports the
competitions closed to date rather than a final standing.}
\label{tab:mle-medals}
\end{table}

The table shows one runtime reaching competitive outcomes across heterogeneous
tasks. The remainder of this section
analyzes the SWE-Bench Pro row because it provides a long sequential trajectory and
Reviewer decisions.

\subsection{Knowing When to Stop: Verification Routing and Non-Termination}

Objective revision is admissible only if the runtime can withhold completion. This
subsection measures where independent verification is spent and where the runtime
declines to declare a task finished.

Of \ReviewerTasks{} tasks, \ReviewerInvoked{} (\ReviewerInvokedPercent\%) invoke
an independent Reviewer, while \ReviewerSkipped{} (\ReviewerSkippedPercent\%) use
Engineer self-review. The Reviewer requests another implementation round on
\ReviewerRevisionRequested{} tasks. After revision, \ReviewerVerifierRecovered{}
pass the official verifier and \ReviewerStrictRescues{} complete the strict
\code{continue}$\rightarrow$revision$\rightarrow$\code{done} loop. A further
\ReviewerFirstBlocked{} tasks receive a \code{blocked} verdict, recording that the
runtime cannot complete them rather than emitting an unsupported completion.
Refusing to stop early and refusing to claim success are two expressions of the same
gate.

Reviewer-routed tasks consume \ReviewerTokenRatio$\times$ as many solve input
tokens and \ReviewerTimeRatio$\times$ as much active time as self-reviewed tasks on
average, indicating that the routing policy selects a harder workload. The recovery
funnel is therefore more interpretable than a raw group-accuracy comparison.

\begin{figure}[H]
  \centering
  \includegraphics[width=\linewidth]{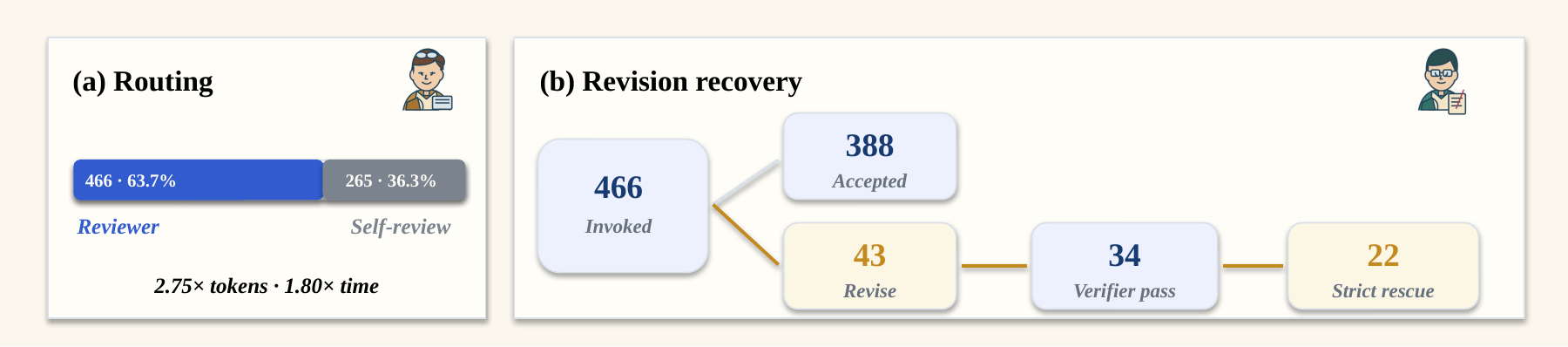}
  \caption{Adaptive Reviewer routing and revision recovery. Panel (a) shows that
  466 of 731 tasks invoke an independent Reviewer, while 265 use Engineer
  self-review; routed tasks consume 2.75$\times$ the solve input Tokens and
  1.80$\times$ the active time on average. Panel (b) follows the routed workload:
  388 tasks are accepted on first review, 43 receive a revision request, 34 of
  those later pass the official verifier, and 22 complete the strict
  review-loop rescue. Routing is task-dependent; a randomized-routing run would
  additionally isolate the causal effect.}
  \label{fig:reviewer-mechanism}
\end{figure}

Figure~\ref{fig:reviewer-mechanism} separates the routing population from the
revision-recovery path. The accepted and revise branches partition outcomes after
Reviewer invocation together with 35 blocked cases; verifier pass and strict rescue
are nested recovery milestones among the 43 revision-requested tasks.

\subsection{Cost of the Mechanism as State Accumulates}

\begin{figure}[H]
  \centering
  \includegraphics[width=\linewidth]{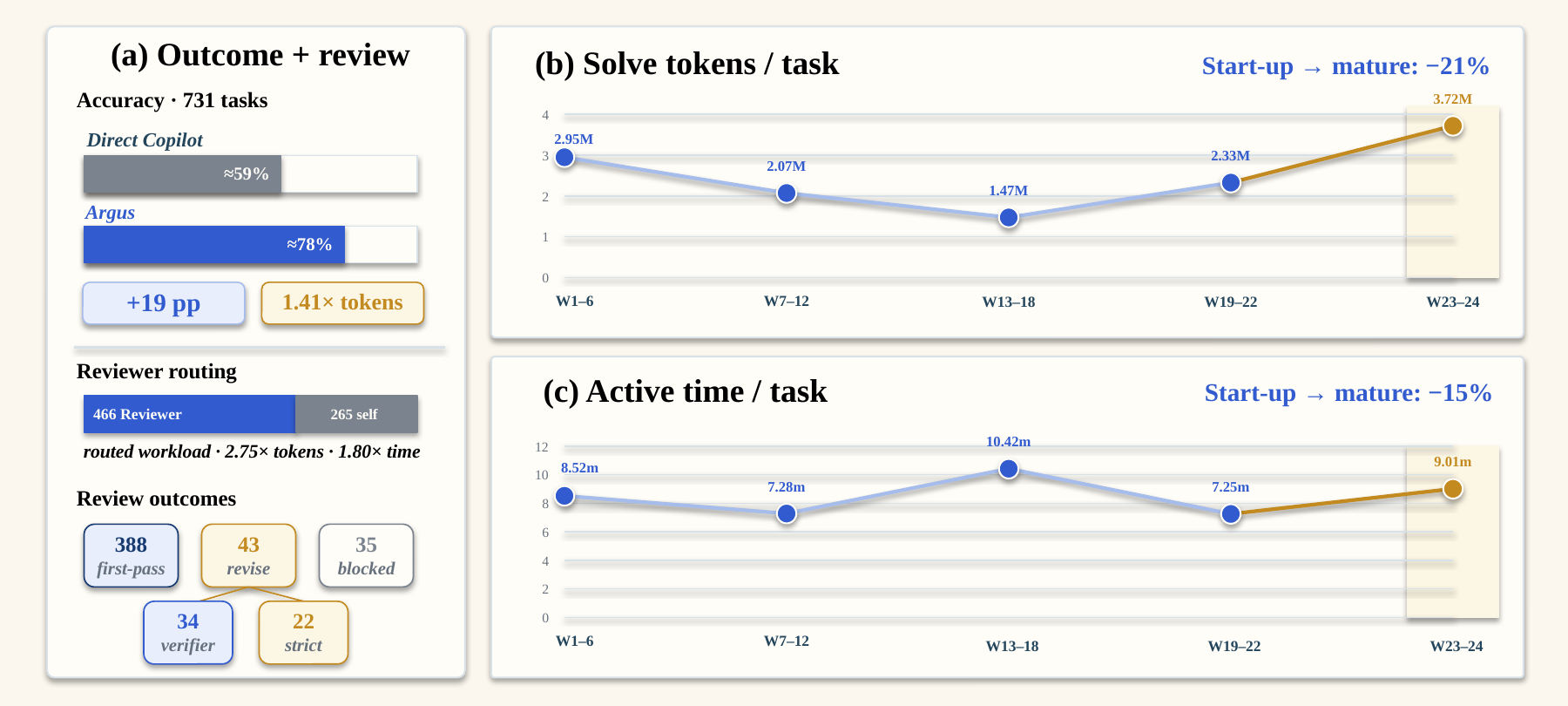}
  \caption{Outcome, review, and longitudinal efficiency in the 731-task SWE-Bench
  Pro run. Panel (a) reports the full-suite accuracy and aggregate-Token comparison,
  adaptive Reviewer routing, and observed review outcomes. Panels (b) and (c)
  aggregate \systemname solve input Tokens and active workflow time over
  approximately six-Wave windows using task-weighted means. Relative to W1--6,
  W19--22 uses 21\% fewer solve input Tokens and 15\% less active time per task.
  Two incomplete Waves are omitted, while W23--24 remains visible as late
  difficult-task stress. Copilot per-Wave resource traces were not retained; the
  window comparison characterizes the operating profile over this sequence, and a
  controlled replay would isolate the learning effect.}
  \label{fig:swebench-evolution}
\end{figure}

Figure~\ref{fig:swebench-evolution} shows the \systemname-only longitudinal
trajectory. Here, self-evolution refers to changes in persistent runtime state, not
to online training of the underlying model. Across Waves, accepted trajectories can
update repository knowledge, reusable Skills, verification guidance, task
definitions, and routing policy. These objects alter what later missions retrieve,
which checks they run, and where they spend independent review.

Mature operation W19--22 uses \SWEProTokenReduction\% fewer solve input Tokens and
\SWEProTimeReduction\% less active workflow time per task than startup W1--6. This
pattern is consistent with accumulated state reducing repeated inspection,
rediscovery, or failed execution. The lowest-Token window, W13--18, reaches a
\SWEProBestTokenReduction\% reduction from startup but requires more active time,
showing that Token efficiency and execution latency do not move together. The
final difficult-task window rebounds in both metrics.

The trajectories show a lower-cost mature operating window together with visible
task-composition and late-wave difficulty effects. The rebound in W23--24 prevents
the startup-to-mature comparison from being read as monotonic improvement. Because
the task sequence is not replayed against a frozen runtime state, the evidence
characterizes system-state accumulation over this sequence; attributing the
reduction to an individual memory or Skill update calls for a matched frozen-state
replay.

\subsection{Measured Process Quantities}

The process theory is instantiated from two complementary traces. In the
mathematical campaign, direct reasoning, verification, and action each account for
approximately 56\% of their respective attributed units. In SWE-Bench Pro, Reviewer
revision requests lead to 79.1\% official-verifier recovery and 51.2\% strict
review-loop rescue, while the mature window uses 21\% fewer solve input Tokens and
15\% less active time than startup. Appendix~\ref{app:formula-instantiation} gives
the complete substitutions for the density, reuse, yield, and compression
quantities.

The public research inventory additionally contains 41 de-duplicated artifacts
across six programs, illustrating the range of research programs executed by the
same runtime.

\subsection{Objective Revision in a Mathematical Campaign}
\label{sec:vertical-trace}

\begin{figure}[H]
  \centering
  \includegraphics[width=\linewidth]{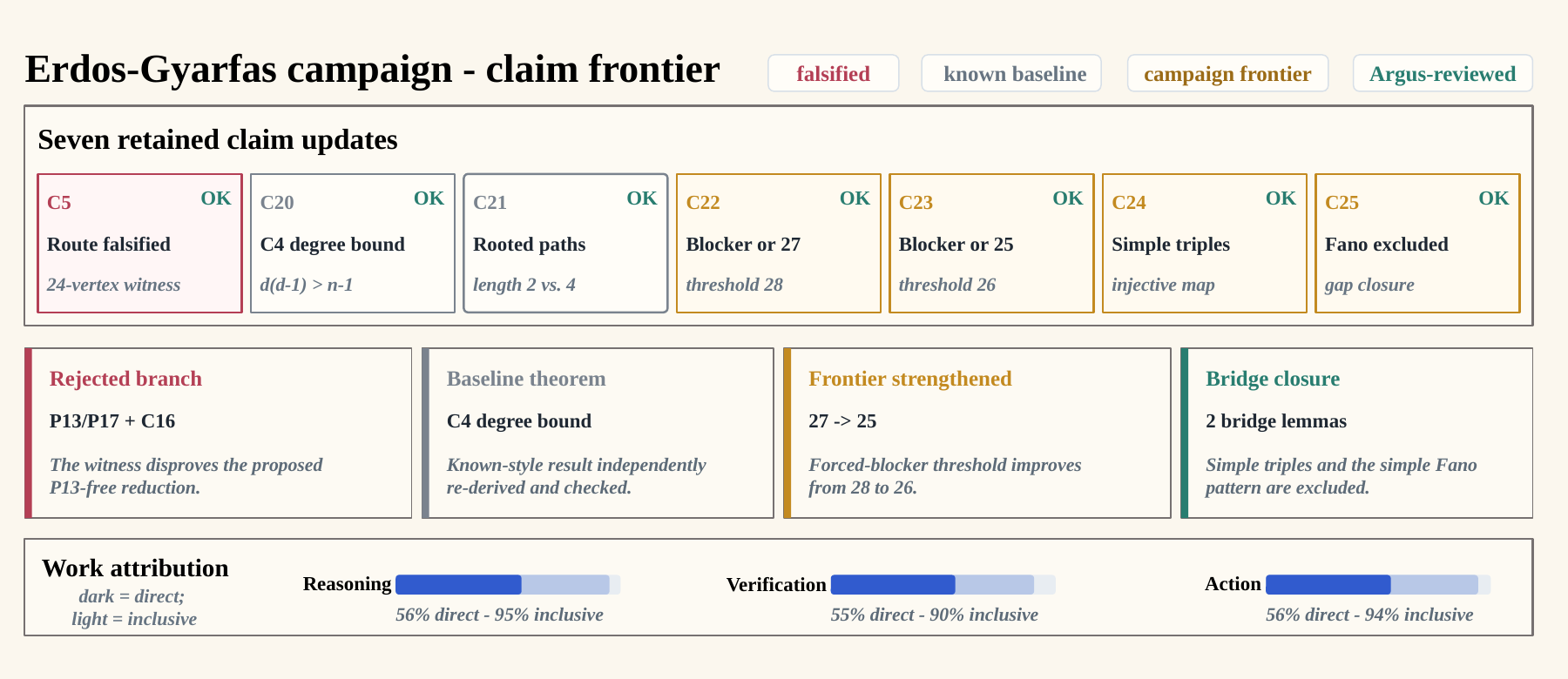}
  \caption{Reviewed claim frontier in the representative mathematical campaign.
  Seven retained updates progress from a falsified route (C5), through two
  independently re-derived baselines (C20--C21), to four campaign-frontier results
  (C22--C25), including the 27-to-25 strengthening and two bridge lemmas. The
  middle cards expose the corresponding artifacts, while the bottom bars separate
  direct frontier-changing work from an inclusive measure that also credits
  validation and state consolidation. Check marks indicate internal \systemname
  review, not external peer review or a novelty claim.}
  \label{fig:erdos-vertical-trace}
\end{figure}

Figure~\ref{fig:erdos-vertical-trace} follows the retained mathematical claim
frontier inside one real campaign. The case begins with an open-ended research objective.
The Manager preserves that goal through budget and process
failures; the Planner turns the current reviewed state into bounded work; the
Engineer retrieves sources, runs tools, and writes research artifacts; and the
Reviewer decides whether the work is accepted, revised, or blocked before it can
enter durable state.

The trace makes the work allocation concrete. The Manager recovers the campaign
and owns stage transitions. The Planner first selects a cheap falsification test,
then proposes changing the success contract from ``collect observations'' to
``produce a proof''; the Manager admits that refinement,
and later requires new missions to advance the retained result rather than restart
from an easier fact. The Engineer performs the domain work---source retrieval,
executable checks, proof writing, and result packaging. The Reviewer rejects an
overstated route, requests the missing checks, and certifies the bounded theorem
recorded by the artifacts.

In the reported trace, the campaign retains one falsified route and six proof-backed
deltas, including one stricter bound and two bridge results. Accepted work changes
the next Planner input, rejected work remains available as a dead branch, and open
blockers become explicit future tasks. The trace therefore captures a continuous
research path rather than a sequence of isolated answers.

\subsection{Efficiency Attribution in the Mathematical Case}

We apply Equation~\ref{eq:operational-efficiency} to the theorem-production
window covering Rounds 12--17. The analysis contains 18 bounded missions: six changed
the theorem frontier, four directly checked or re-certified mathematical results,
seven consolidated review state and documentation, and one overlap-analysis mission was interrupted
without an admitted result. Table~\ref{tab:erdos-efficiency} uses mission purpose as
a coarse attribution rule for process allocation.

\begin{table}[H]
\centering
\footnotesize
\renewcommand{\arraystretch}{1.12}
\begin{tabularx}{\linewidth}{@{}p{2.65cm}rrrX@{}}
\toprule
\rowcolor{papersand}
\textbf{Efficiency view} & \textbf{Direct} & \textbf{Auxiliary} & \textbf{Failed/no-op} & \textbf{Unit and interpretation} \\
\midrule
Reasoning & 56.0\% & 39.1\% & 4.9\% & Engineer reasoning-output Tokens; direct means a frontier-changing proof mission. \\
Verification & 55.4\% & 35.0\% & 9.6\% & Reviewer reasoning-output Tokens; direct means proof comparison or re-certification. \\
Action & 55.6\% & 38.9\% & 5.6\% & Mission count; direct comprises six frontier missions and four verification missions. \\
\bottomrule
\end{tabularx}
\caption{Efficiency attribution for the representative mathematical case.
Auxiliary work includes result packaging, checkpoint updates, and documentation
consolidation.}
\label{tab:erdos-efficiency}
\end{table}

The normalized interpretation is concrete. Of every 100 Engineer
reasoning-output Tokens, approximately 56 occurred in frontier-changing proof
missions, 39 in verification or state-maintenance work, and 5 in the interrupted
branch. Of every 100 Reviewer reasoning-output Tokens, approximately 55 directly
checked a proof package, 35 maintained formal review state, and 10 were
spent in the interrupted analysis. The strict action score is $10/18=55.6\%$; if the
seven successful auxiliary missions are also credited, the inclusive score is
$17/18=94.4\%$. Cost weighting makes the interruption more visible: the strict and
inclusive action scores become 56.1\% and 89.7\%, respectively.

The trace also clarifies what the numerator should reward. The early witness check
that killed a proposed reduction counts as an efficient action because it pruned a
false route before a long proof attempt. In Round 15, selecting one internal triple,
deriving the degree bounds, and closing the counting inequality are direct reasoning;
rewriting the result into the structured claim record is auxiliary work. Reviewer
inspection of those steps is direct verification, whereas updating the working
checkpoint is coordination overhead. This separation prevents activity volume from being mistaken for research
progress while preserving necessary coordination as a visible cost.

\subsection{Objective Revision at Campaign Scale}
\label{sec:paper-case-study}

The benchmark suite evaluates task endpoints, while the mathematical trace follows
one research objective in depth. A complementary question is whether the same
runtime can carry multiple research programs through the complete paper-production
lifecycle. We therefore reconstruct six projects spanning evaluation reliability,
vision--language matching, test-time adaptation, GUI agents, multimodal
hallucination, and model quantization. The initial objectives and compute
environments were provided externally; the retained Argus records identify the
Manager, Planner, Engineer, and Reviewer as the actors responsible for Stage
transitions, experiments, manuscript construction, review, and submission checks.

\begin{figure}[H]
  \centering
  \includegraphics[width=\linewidth]{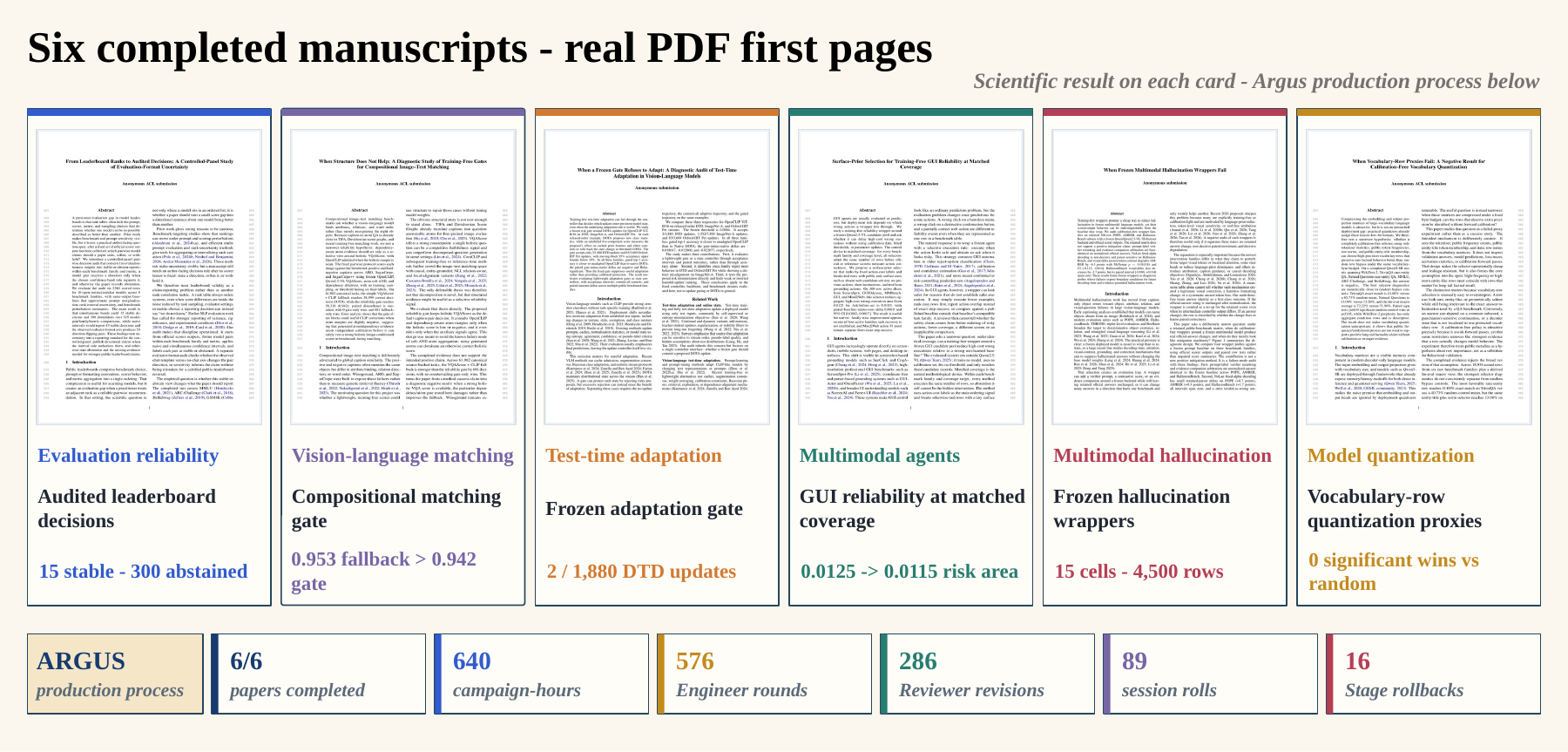}
  \caption{Scientific outcomes and production load across six paper campaigns.
  Each card pairs the final manuscript's first page with its domain, scoped research
  question, and principal task-native result, including diagnostic and negative
  findings. The bottom strip reports the shared production process: six completed
  pipelines, 640 aggregate campaign-hours (rounded from 639.55), 576 Engineer rounds, 286 Reviewer
  revisions, 89 session rolls, and 16 Stage rollbacks. Manuscript completion does
  not imply venue acceptance, novelty, or external peer review.}
  \label{fig:paper-case-study}
\end{figure}

Figure~\ref{fig:paper-case-study} summarizes the portfolio. All
\PaperCaseCompleted{} of \PaperCasePapers{} canonical pipelines reach the final
submission Stage. In aggregate, they span \PaperCaseCampaignHours{} campaign-hours,
\PaperCaseMissions{} bounded missions, \PaperCaseRounds{} Engineer rounds,
\PaperCaseContinues{} Reviewer revision verdicts, \PaperCaseSessionRolls{} session
rolls, and \PaperCaseRollbacks{} Manager rollbacks. Campaign-hours are summed across
projects and are not calendar time because several projects overlap.

The trajectories are inconsistent with a one-pass account of autonomous paper
writing. Each manuscript states a scoped research question and a task-native
outcome rather than merely presenting a generated document. The portfolio includes
uncertainty-aware leaderboard reporting, a failed compositional gate, an
over-restrictive adaptation controller, a narrow GUI reliability gain, a
failure-mode audit of multimodal wrappers, and a negative result for static
quantization proxies. Several projects therefore become diagnostic or negative
results after their original positive hypothesis fails. Verification-gated state makes
that scientific pivot explicit instead of silently replacing the failed branch.

\begin{figure}[H]
  \centering
  \includegraphics[width=\linewidth]{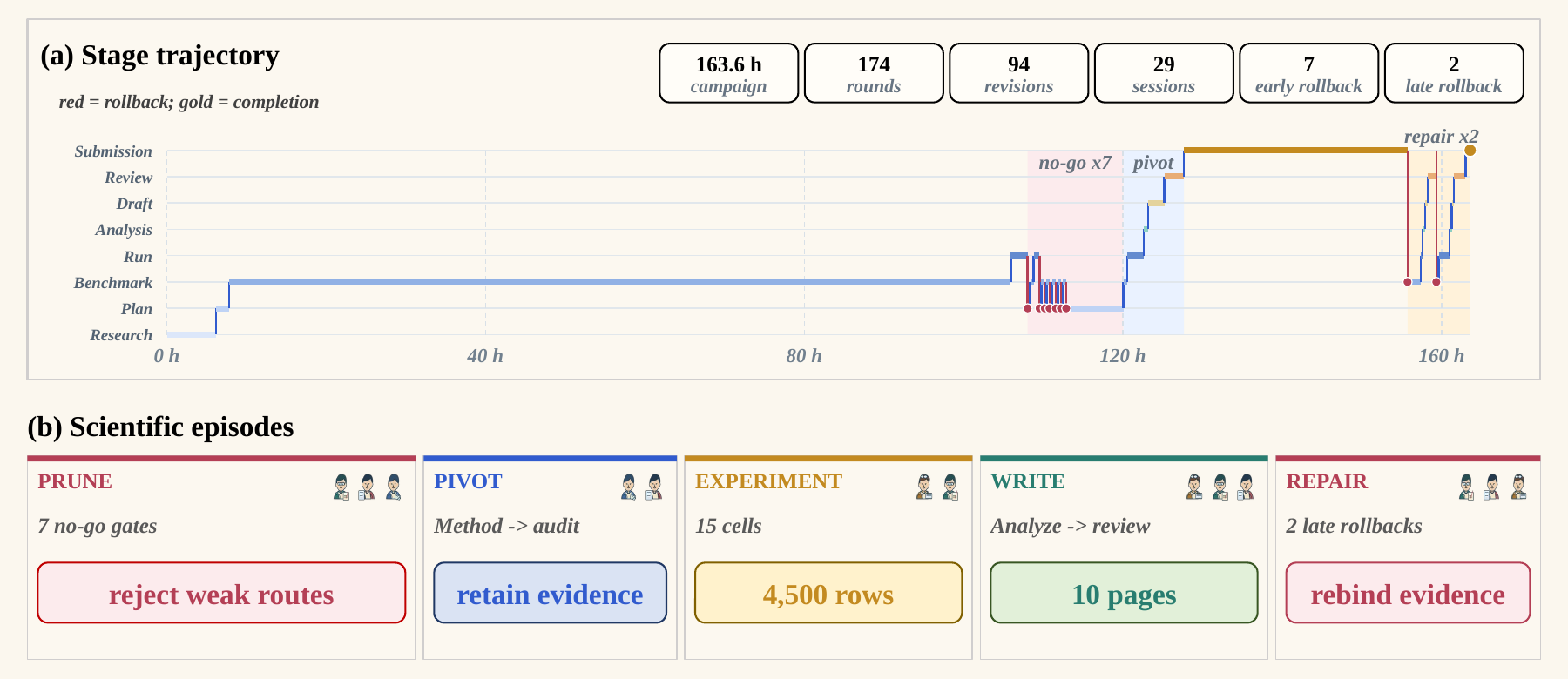}
  \caption{Representative 163.6-hour paper-production trajectory. Panel (a) plots
  the Manager-controlled Stage against campaign time: seven early no-go decisions
  prune weak method routes, the project pivots from a positive method claim to an
  audit, and two late rollbacks repair the submission evidence. Panel (b) reduces
  the trace to five scientific episodes---prune, pivot, experiment, write, and
  repair---while retaining the 15-cell, 4,500-row experiment and the final
  10-page manuscript.}
  \label{fig:paper-case-trajectory}
\end{figure}

Figure~\ref{fig:paper-case-trajectory} resolves the multimodal-hallucination
campaign in greater detail. During a dense 12-hour search window, seven rollback
decisions reject baseline coverage or proposed mechanisms that are unreproduced,
base-identical, or missing their preregistered signal. Planner then proposes changing
the claim from a positive mitigation method to a diagnostic negative-results audit,
and Manager admission makes the refinement authoritative.
Engineer completes a five-method by three-benchmark matrix with 4,500
official-scored rows; Reviewer binds the resulting no-op and degradation claims to
those outputs. Two later submission checks return the project to benchmark to
repair scorer provenance and GPU telemetry before the final 10-page AAAI-formatted
package completes the pipeline.

The process layer is similarly nontrivial. Even the shortest case requires 28
Engineer rounds, 19 Reviewer revisions, and 13 session rolls over 20.8 hours. The
two AAAI-formatted cases finish as 10-page anonymous submissions, while the four
ACL-formatted cases contain 10--13 pages. Across all six projects, the retained
academic, layout, and infrastructure histories contain
\PaperCaseReviewSnapshots{} automated review snapshots. These are process records
rather than external assessments of paper quality.

The case study also exposes a consistency failure. The compositional-matching
pipeline reaches a complete submission Stage and produces its final PDF, while the
last stored assurance object remains \code{BLOCKED} on a Manager-stage authority
check. This discrepancy does not change the paper artifact, but it shows that
derived certification state can lag the canonical pipeline state. Public release
automation should reconcile those two views before presenting assurance as a final
verdict.

\subsection{A Chip the Runtime Designed: ACE-2}
\label{sec:silicon-vertical}

The campaigns above produce software, proofs, and manuscripts. ACE-2 asks whether
the same runtime holds when the artifact is hardware, where an unverified claim
survives exactly as long as it takes someone to run the tool that contradicts it.
\systemname specified ACE-2, wrote its RTL, built its verification environment,
and drove it through mapped synthesis and static timing without a human author of
record for any of those artifacts. The result is an inference accelerator that
executes Qwen2.5-0.5B end to end in W4A8, certified against a fixed scope.

\paragraph{Functional closure.} The accepted design runs the full 24-layer
Qwen2.5-0.5B W4A8 command integration. Layer 0 matches the reference on all 18
ordered operators exactly, and the two-token runtime completes
13{,}914/13{,}914 commands over 1{,}240{,}410{,}384 simulator cycles with
generated token identifiers $[0,0]$ and no first failure recorded. The runtime
package, command log, and progress journal are each bound by content hash, so the
functional claim resolves to specific bytes rather than to a summary.

\paragraph{Physical closure.} Canonical SKY130 HD mapped synthesis and OpenSTA at
TT 25\,\textdegree C / 1.80\,V report 62{,}283 cells and 0.614\,mm$^2$ of
non-SRAM area against an operator-set cap of 2.0\,mm$^2$, with $+0.6966$\,ns
detailed setup slack, 0.00\,ns worst negative slack, 0.00\,ns total negative
slack, and a 10.000\,ns clock period. Both operator-owned targets---the area cap
and the 100\,MHz floor---pass without relaxation, and the passing packet is the
sole exactly-once canonical run rather than the best of several attempts.

\paragraph{What the gate produced.} The certification is dated 2026-08-04 and was
issued by a Reviewer that did not perform the work, bound to one RTL tree hash.
Each accepted RTL repair carries its own reviewer binding together with the
originating and resulting tree hashes, so the certified design is reachable from
the audit trail rather than asserted alongside it. The more informative artifact,
for the argument of this report, is what the certificate refuses to say. It
enumerates its own exclusions: no routed timing, no power signoff, no DRC/LVS, no
GDS or tapeout, no silicon validation, no generation beyond two tokens, no
external deployment interfaces, and no FPGA prototype. That list was produced by
the same gate that admitted the positive claims. An unbounded runtime would have
reported a chip; this one reported a chip and the exact perimeter of the evidence
supporting it.

ACE-2 therefore extends the pattern of Sections~\ref{sec:vertical-trace}
and~\ref{sec:paper-case-study} into a domain with an unusually unforgiving
verifier. The functional and physical claims are narrow by construction, and the
scope conditions are part of the output rather than a caveat added afterward.

\subsection{Simplifying a Published Method: Materials Generation}
\label{sec:materials-vertical}

The verticals above are evaluated by a proof checker, a venue format, and a
synthesis toolchain. This one is evaluated by an external chemistry validator that
the runtime does not own, against two published models it did not write. The
campaign targets metal--organic framework (MOF) generation, and its outcome is the
sharpest instance in this report of the pattern Section~\ref{sec:method} describes:
the admitted result is \emph{simpler} than the method it replaces, and it was
admitted only after a confound in the original comparison was removed.

\paragraph{Adding a control the base model lacked.} MOFFlow-2 generates
building-block sequences with an autoregressive language model and then places them
by flow matching. Its published conditioning accepts a single continuous property
and cannot address chemical identity. \systemname added metal element, node
nuclearity, and ligand family as three discrete condition tokens read through
cross-attention, each with its own \code{ANY} token and independent condition
dropout, so one model serves unconditional, partial, and joint requests. Continuing
to train the full decoder raised validation loss, so the language model was frozen
and only the condition encoder and six cross-attention modules were updated:
12.69M trainable parameters of 75.8M, or 16.7\%. Over 3{,}300 balanced condition
requests weighted toward rare chemistry, adherence reaches 92.5\% on metal
(17.4\% under a permuted-condition control), 100.0\% on nuclearity (24.3\%), and
74.5\% on ligand family (39.0\%). The effect concentrates in the tail the corpus
under-represents: nickel rises from a 2.1\% corpus base rate to 83.7\%, trinuclear
nodes from 3.9\% to 100.0\%, and pyrazolate ligands from 2.0\% to 46.0\%. Unconditional
structure-level validity also improves, from 30.61\% for the local MOFFlow-2
reproduction to 37.12\%. The control is not free: at guidance $w{=}2$ structural
validity falls to 17.18\%, and the campaign records that trade-off rather than
reporting only the adherence numbers.

\paragraph{Replacing the score, then questioning the method.} A failure census over
19{,}483 paired structures found that the published steering potential penalizes
atom overlap, while the dominant failures of the all-atom model are the opposite:
under-coordination, fragments, and floating components. \systemname built a scoring
function aligned to the external \code{MOFChecker} criteria instead, with radii
derived from covalent and van der Waals tables rather than tuned by hand. Against
the external validator it separates passing from failing structures at
AUC 0.833, versus 0.594 for the overlap-only potential.

That gain then exposed a measurement problem. Enabling the published Feynman--Kac
steering also switches the integrator from ODE to SDE, so the reported improvement
mixes two changes. Separating them attributes $+1.92$ points to the integrator alone,
which does not reach significance ($p{=}0.238$), and $+9.00$ points to the score
inside a fixed SDE ($p{=}2.0\times10^{-9}$). With that confound removed, the runtime
tested whether the particle interaction was doing any work at all, by holding model,
integrator, score, forward-pass budget, and the same 989 crystals fixed and simply
generating $K$ independent trajectories and keeping the best-scoring final structure.

\begin{table}[H]
\centering
\small
\renewcommand{\arraystretch}{1.16}
\begin{tabularx}{\linewidth}{@{}Xp{2.5cm}p{2.5cm}p{1.9cm}@{}}
\toprule
\rowcolor{papersand}
\textbf{Forward passes} & \textbf{Feynman--Kac} & \textbf{\systemname best-of-$K$} & \textbf{Difference} \\
\midrule
$K=4$ & --- & 53.39\% & --- \\
$K=8$ & 52.38\% & \textbf{55.21\%} & \textbf{$+2.83$} \\
$K=16$ & 57.94\% & \textbf{59.76\%} & $+1.82$ \\
$K=32$ & 59.96\% & \textbf{61.38\%} & $+1.42$ \\
\bottomrule
\end{tabularx}
\caption{External \code{MOFChecker} pass rate on 989 paired crystals at matched
forward-pass budget, model, integrator, and score. Single-sample SDE decoding passes
43.38\%. Independent generation followed by best-of-$K$ selection beats the published
Feynman--Kac steering at every budget; the $K{=}8$ difference survives a paired test
($p{=}0.0486$). Best-of-4 at 53.39\% already exceeds Feynman--Kac at $K{=}8$ while
spending half the compute.}
\label{tab:mof-bestofk}
\end{table}

Table~\ref{tab:mof-bestofk} reports the result. The mechanism is visible in the
particle statistics: at $\lambda{=}2$ the median effective sample size is
approximately one, and 72.1\% of crystals collapse to a single particle after
resampling, after which the remaining SDE noise does not restore diversity. The
published method pays for $K$ forward passes and then discards most of the
exploration it bought. Keeping the trajectories independent costs nothing extra and
retains it.

\paragraph{Where the campaign stops.} This is a research candidate, not a
state-of-the-art claim, and the runtime records why. The 989-crystal subset is the
same data on which the score was designed, so the numbers above are not an
independent holdout. The full 19{,}483-structure leaderboard has not been run. The
matched-compute MOFFlow-2 comparison was interrupted by a parameter error that
enumerated all 19{,}792 test structures instead of the intended 1{,}500, and its
export is incomplete. Conditional sequence generation and best-of-$K$ structure
prediction have not been joined into one end-to-end system. No relaxation, machine-learned
interatomic potential, or DFT check has been run, so every number here is scored by
the same validator the selection optimizes against, which is exactly the failure mode
Section~\ref{sec:limitations} names for verifier-shaped objectives. The campaign
therefore holds at a candidate version rather than declaring a release, and the
outstanding items are enumerated rather than deferred silently.

What this vertical contributes to the argument of the report is not the pass rate.
It is that a runtime given a published method, a fixed compute budget, and an
external validator arrived at a smaller method than the one it started from, and
that the step which made this admissible was removing a confound rather than adding
a component.

\section{Analysis and Discussion}
\label{sec:discussion}

\subsection{System-Level Performance}

The seven-benchmark suite shows one runtime operating across software repair,
GPU-kernel optimization, model training, research tasks, and data synthesis; a
separate mathematical vertical trace examines proof-oriented research depth. On
SWE-Bench Pro, \systemname reaches approximately 78\%
accuracy compared with 59\% for Direct Copilot at 1.41$\times$ aggregate Tokens.
The comparison pairs additional computation for planning, execution, and review
with a 19-point accuracy difference.

The longitudinal run shows how this cost changes with accumulated state. Mature
W19--22 uses 21\% fewer solve input Tokens and 15\% less active time per task than
startup W1--6. The curve remains non-monotone: W13--18 combines low Token use with
longer execution, while W23--24 contains a late concentration of difficult tasks. The operating profile changes
over time as shared state expands, while task-dependent variation remains visible.

\subsection{Reviewer as Selective Error Correction}

The Reviewer requests revision on 43 SWE-Bench Pro tasks. After another Engineer
round, 34 pass the official verifier and 22 complete the stricter
\code{continue}$\rightarrow$revision$\rightarrow$\code{done} loop. These trajectories
show the Reviewer functioning as an error-correction stage rather than a terminal
commentary layer.

Reviewer routing is adaptive: Reviewer-routed tasks use more Tokens and active time
than self-reviewed tasks. The observed routing concentrates independent review on
the more resource-intensive part of the workload.

\subsection{Autonomous Research Production}

The six-paper case study extends the endpoint results into a complete research
lifecycle. All six canonical pipelines reach submission completion across six
domains and two venue formats, but none follows a one-pass path. Reviewer revision
verdicts, Stage rollbacks, and session rolls are common rather than exceptional.
The final manuscripts therefore demonstrate persistence across research framing,
experimentation, writing, review, and packaging---not merely the ability to emit
paper-like prose.

The multimodal-hallucination trace illustrates the control mechanism. Reviewer
gates reject seven weak method routes before they are scaled into an unsupported
positive claim; Planner proposes reframing the accumulated failures as a diagnostic study;
Engineer then completes the 15-cell canonical matrix; and Manager accepts two late
rollbacks when submission checks expose scorer-provenance and telemetry defects.
The recorded sequence is consistent with role separation redirecting the scientific
trajectory: failure is
retained as information, while only reviewed state controls the next Stage.

The case also separates artifact completion from process certification. Five final
assurance snapshots pass; the compositional-matching project retains a stale blocked
assurance snapshot after its canonical pipeline and final PDF complete. This is a
useful systems distinction: derived quality metadata must be reconciled with the
authoritative Stage state before it can serve as a public certificate.

\subsection{Persistent State and Verification-Gated Runtime Self-Evolution}

The long-lived object in \systemname is the campaign state rather than a provider
transcript. Accepted results, failed routes, tools, skills, and current blockers
remain available after session and process restarts. Equation~\ref{eq:process-data}
formalizes the information advantage of this record, while
Equation~\ref{eq:process-compression} explains why the runtime keeps both a complete
event history and a bounded working checkpoint.

This mechanism gives concrete meaning to \emph{verification-gated fixed-model runtime
self-evolution}. The base model remains unchanged, but an admitted mission can
change the starting point, available procedures, verification obligations, or
routing policy of later work. The admission gate binds each candidate to artifacts,
task-native evidence, and an authorized commit; independent Reviewer judgment is one
gate path rather than a requirement for every low-risk update. Rejected branches
also participate when they are retained as verified exclusions: later missions can
avoid repeating the branch and cite its failed evidence when proposing a pivot.
Evolution therefore occurs in the runtime's admitted state and control policy, not
in model weights or an unconstrained summary of prior conversation.

The two longitudinal views expose complementary parts of this claim. The SWE-Bench
Pro trajectory measures how solve Tokens and active time vary as shared state
accumulates, while the mathematical campaign shows direct semantic reuse: later
missions consume earlier definitions, bounds, rejected routes, and bridge lemmas.
Neither trace isolates a single update mechanism. A matched experiment that replays
the same tasks with frozen versus accumulated runtime state would be required to
attribute a causal gain to memory, Skills, verification, or routing separately.

\subsection{Endogenous Harnessing: When a Plan Becomes a Constraint}
\label{sec:endogenous-harness}

A harness is supposed to bound what a system may do. The instructive failure is when
a system manufactures its own harness out of a decision it made earlier, and then
obeys it. We call this \emph{endogenous harnessing}.

The mechanism follows directly from the two levels in Algorithm~1. A Planner proposes
a strategy under the information available at planning time. That proposal is written
into a mission, and from the first round onward the Engineer treats it as the task and
the Reviewer treats it as the standard the work is judged against. What began as one
agent's hypothesis has become an external constraint on every agent that comes after,
including agents that now know more. The system stops behaving like a researcher
continuously reconsidering the best route and starts behaving like a researcher bound
by a note they wrote before the experiment ran.

A recorded verification trace shows the shape. The Planner required a
\code{skip-zero} candidate. The Engineer satisfied that local objective. The Reviewer
then identified a no-gap validator alternative that would have discharged the same
obligation more directly. The Reviewer, however, is scoped to the current round: it
can reject, request revision, or escalate, but it cannot redefine the mission that
made the weaker route mandatory. The better judgement arrived, was recorded, and did
not change the task. The earlier, less informed decision won on authority rather than
on evidence.\footnote{This trace is drawn from internal system-verification work. We
report it as a described sequence rather than a public reproducible artifact.}

Planning is not the problem here, and neither is bounding. The problem is category
confusion between two things a plan can be. A plan can be a \emph{contract}, which is
binding and expensive to change, or a \emph{falsifiable hypothesis}, which is
supposed to be discarded the moment the evidence turns. \systemname currently records
both in the same place, so a technical strategy inherits the immovability that should
belong only to a safety boundary. The correct division is sharper than the one the
runtime implements: frozen authority, prohibitions on granting new trust, and
irreversibility limits are hard harness and should resist revision; the choice of
route, validator, or representation is a technical bet and should be revisable by any
role that produces evidence against it.

The implemented contract already encodes half of this distinction. As
Section~\ref{sec:method} describes, semantic clauses move freely while precise
clauses and the objective require explicit confirmation. What is missing is that a
mission's technical strategy is filed on the immovable side of that line rather than
the negotiable one, and that the Reviewer, which is the role most likely to hold
disconfirming evidence, has no channel to move it.

This reframes where the current bottleneck sits. The failures above are not reasoning
failures. The system found the counterexample, identified the validator alternative,
and understood the cascading invariant. It could not act on any of them in time
because the workflow gave later, better-informed roles no authority to revise the task
boundary. The binding constraint on this class of system is methodological rather than
cognitive, which is also why the remedy is a change in authority routing rather than a
stronger model.

\subsection{From Runtime Learning to Model Training}

The same trajectories can be transformed into supervised examples, preference
pairs, and reinforcement-learning transitions. AgentInstruct, Agent-FLAN, and
Agent Lightning provide complementary mechanisms for this conversion
~\citep{mitra2024agentinstruct,chen2024agentflan,luo2025agentlightning}.
\systemname contributes a structured source stream linking objectives, tool actions,
measurements, revisions, and retained-state updates. A future training stage can use
this stream to internalize recurring planning and verification patterns while the
runtime continues to provide long-horizon coordination.

\subsection{Vertical Research in Mathematics}

The Erd\H{o}s--Gy\'arf\'as campaign exposes the four-role organization at research
depth. The Manager preserves the objective and stage, the Planner converts the
current frontier into bounded theorem tasks, the Engineer produces proofs and
computational checks, and the Reviewer controls admission into durable state.

The campaign retains one falsified route and six accepted theorem-frontier updates.
Later missions improve a bound and close two graph-realizability gaps by starting
from the reviewed ledger rather than restarting from the original conjecture. This
is the same runtime mechanism seen in software repair, applied to a domain where
progress is expressed through definitions, lemmas, counterexamples, and proofs.

\subsection{Verticals Whose Verifier the Runtime Does Not Own}

The chip and materials campaigns (Sections~\ref{sec:silicon-vertical}
and~\ref{sec:materials-vertical}) sharpen the same argument under a stricter
condition: in both, the deciding evaluator is external. A static-timing engine and
an independent chemistry validator will contradict an unsupported claim the moment
they are run, so the gate cannot be satisfied by narrative.

Their outcomes are informative in opposite directions, and both are consistent with
the position of Section~\ref{sec:problem}. ACE-2 closes a scope and then publishes
the perimeter of that scope as part of the result, so the exclusions are an output of
the gate rather than a caveat appended to it. The materials campaign instead reverses
a design assumption: after separating an integrator change from a scoring change, the
runtime found that the published particle-interaction step was discarding the
exploration it paid for, and that keeping $K$ trajectories independent scored higher
at matched compute. The admitted method is therefore smaller than the one it replaced.
Neither outcome required a new component, and neither was reported as a
state-of-the-art claim, because in one case the evidence stops at mapped synthesis and
in the other it stops at the same validator the selection optimizes against.

\subsection{Connecting Theory and Results}

Table~\ref{tab:theory-results} connects the process-to-capability model to the
measurements reported in Sections~\ref{sec:results} and~\ref{sec:vertical-trace}.

\begin{table}[H]
\centering
\scriptsize
\renewcommand{\arraystretch}{1.12}
\begin{tabularx}{\linewidth}{@{}p{2.4cm}p{3.55cm}p{2.55cm}X@{}}
\toprule
\rowcolor{papersand}
\textbf{Theory component} & \textbf{Measurement} & \textbf{Observed result} & \textbf{Interpretation} \\
\midrule
Process-data dominance & Mathematical trajectory & One dead route and six dependent theorem updates & The retained process changes the starting state of later missions. \\
Review gate & 43 Reviewer revision requests & 34 verifier recoveries; 22 strict rescues & Independent review redirects initially unacceptable work into another implementation round. \\
Reuse value $G_L$ & Startup W1--6 versus mature W19--22 & 21\% fewer solve input Tokens; 15\% less active time & Later operation reaches a lower resource regime while shared state expands. \\
Recovery under persistent state & Paper-production trajectory & Seven early no-go rollbacks, two late submission repairs, and 29 session rolls & The campaign changes hypothesis and repairs evidence without losing accepted results or restarting the manuscript. \\
Token conversion & Rounds 12--17 role attribution & 56.0\% reasoning, 55.4\% verification, and 55.6\% action direct shares & The research trace separates frontier work, coordination, and interrupted work. \\
\bottomrule
\end{tabularx}
\caption{Connection between the process-to-capability theory and measured system
behavior. Detailed substitutions appear in Appendix~\ref{app:formula-instantiation}.}
\label{tab:theory-results}
\end{table}

\section{Limitations and Future Work}
\label{sec:limitations}

\paragraph{User-guided pivots are not publicly evaluated prospectively.}
Internal system-verification use has produced concrete cases in which stopping or
surfacing a contradiction helped real researchers refine the operational task.
Those cases contain project-specific research details and user interactions that
cannot be disclosed here. They motivate the mechanism but do not constitute a
public prospective study of clarification quality, researcher calibration, or time
to identify the binding constraint. A publishable evaluation requires consented,
prospectively designed collaboration studies with disclosure-safe tasks.

\paragraph{Contract refinement can still fail.}
Evidence and explicit authority reduce goal drift but do not eliminate it. A
Manager or operator can approve a poorly framed tradeoff, a recorded standing
intent can omit a tacit requirement, and the report-level $K_t/X_t$ projection is
distributed across several runtime surfaces rather than one atomic transaction.
Future work should test adversarial refinements, invariant-preservation checks,
disagreement handling, and rollback of harmful contract changes.

\paragraph{Verification is only as sound as its evidence boundary.}
An executable test, formal checker, benchmark, or model Reviewer can encode the
wrong property. The runtime can record and revise a verifier, but that does not make
the verifier correct. Controlled studies should separately seed implementation
bugs, specification bugs, and mismatches between them, then measure whether the
runtime localizes the faulty side rather than merely satisfying the current check.

\paragraph{Attribution and task sequence.}
The SWE-Bench Pro experiment compares the complete \systemname runtime with Direct
Copilot. Reviewer routing is adaptive, the 22 completed Waves (24 Wave IDs with two
incomplete Waves excluded) follow one task order, and
per-Wave Direct-Copilot Token/time records are unavailable. The current results
therefore characterize whole-system behavior over this sequence. Matched
frozen-state runs, randomized task orders, and randomized review routing will
separate the contributions of persistent state, role separation, and review. The
startup-to-mature comparison is observational rather than a causal learning ablation.

\paragraph{Generality across models, domains, and evaluators.}
The seven benchmark arenas use different metrics, hardware, and execution
protocols, while the mathematical analysis follows one vertical campaign. Results
are interpreted in their native units rather than combined into a universal score.
The in-progress GLM-5.2 run on Claude Code is the first evidence that the runtime
transfers across both backbone and execution surface, but it is incomplete and has
no matched Direct baseline, so it bounds nothing yet. Repeated campaigns,
additional backbones, and external domain review will measure how the runtime
transfers across task families and research standards.

\paragraph{ACE-2 is certified for a demonstrated scope, not for silicon.}
The chip result establishes functional closure of a 24-layer, two-token
Qwen2.5-0.5B W4A8 integration and canonical mapped SKY130 synthesis with static
timing. It is not routed timing, power signoff, DRC/LVS, GDS or tapeout, silicon
validation, generation beyond two tokens, or an FPGA prototype, and the
certificate records those exclusions itself. Whether the same runtime reaches
signoff-quality physical design or fabricated silicon is untested.

\paragraph{Metric calibration and model transfer.}
The efficiency analysis labels whole role calls by mission purpose; finer segment
labels would sharpen the reasoning, action, and verification factors. Reviewer
sensitivity and false-acceptance rate require randomized routing with an external
verifier, and the counterfactual reuse value $G_L$ requires paired future tasks with
and without a state update. The runtime-to-model path is currently a training
direction: held-out trajectory splits and scaffold-removal studies will measure how
much of the long-horizon control loop can be internalized by the model.

\paragraph{Scope of the paper-production case study.}
The six paper projects come from one shared research environment, campaign-hours
overlap in calendar time, stored review snapshots are model-generated, and runtime
logging evolved across projects. Within that scope, the case study establishes
end-to-end lifecycle behavior and recovery under review across six domains and two
venue formats, not paper acceptance, scientific novelty, or superiority to human
research teams. It also does not establish a measured zero-touch rate. Recorded
costs in the public data exclude utility calls without a priced event.

\section{Conclusion}
\label{sec:conclusion}

\systemname is a persistent runtime for research that arrives underdefined. Its
premise is that the hard case for self-improving agents is not a harder score to
climb but the absence of a trustworthy score at all. Where supervision is dense,
iteration steers itself; where the objective, the constraints, and the right
measurement are still being discovered, a runtime earns its keep by holding a
campaign together until they are settled, and by handing the questions it cannot
settle to a human expert instead of guessing past them. The runtime should therefore
persist while evidence supports the current route and pivot when verification exposes
a failed approach, hidden constraint, or misspecified objective. Persistent campaign state, four-role authority separation,
and verification-gated updates carry verified progress across model calls, tool runs,
revisions, and restarts. Across seven benchmark arenas, the runtime reaches strong
task-native results; on SWE-Bench Pro it attains approximately 78\% accuracy versus
59\% for Direct Copilot at 1.41$\times$ aggregate Tokens. Mature operation uses
21\% fewer solve input Tokens and 15\% less active time than startup, while Reviewer
intervention produces 34 verifier recoveries and 22 strict review-loop rescues.

The self-evolution claim is deliberately fixed-model and verification-gated. Candidate
memories, skills, procedures, verifiers, routing decisions, and rejected routes alter
future work only after evidence checks and an authorized commit; model parameters do
not change. Thus a failed branch can become reusable progress without allowing an
unverified narrative to rewrite the campaign.

The report-level $K_t/X_t$ model and \code{ManagerAdmit} operator make the
evidence-and-authority requirements for contract refinement explicit; they summarize
distributed runtime surfaces rather than one atomic API. The process-to-capability
theory explains how this runtime converts Token activity
into retained research progress. Typed trajectories contain more decision-relevant
information than final artifacts alone, review separates proposal from admission,
and reuse changes the starting point of future missions. The mathematical
campaign makes this mechanism visible through one rejected route and six accepted
theorem-frontier updates produced by the Manager, Planner, Engineer, and Reviewer.

The six-paper production case provides a second vertical view. Across
\PaperCaseCampaignHours{} aggregate campaign-hours, Argus repeatedly survives
review rejection, Stage rollback, and session replacement while producing two
AAAI- and four ACL-formatted manuscripts. In the representative campaign, seven
rejected method routes become a 4,500-row negative-results study, and two late
submission rollbacks repair the evidence package without resetting the paper. This
illustrates how the runtime can connect hypothesis revision, experiment
execution, and sustained scientific communication, while the stale assurance
snapshot in one project identifies a concrete synchronization failure for future
work.

These trajectories also define a natural interface to model training. Supervised,
preference-based, and reinforcement-learning methods can use the accumulated
objectives, actions, measurements, revisions, and state transitions to internalize
parts of the long-horizon research loop.

\clearpage
\section*{Contributors}
\addcontentsline{toc}{section}{Contributors}

\systemname is built by a team across ten institutions. Within each group below,
names are listed alphabetically by given name.

\vspace{0.4em}
\noindent{\sffamily\bfseries\color{argusdeep}Core Contributors}\par
\vspace{0.15em}
\noindent
Boxiu Li (Shanghai Jiao Tong University; Microsoft),
Zimo Wen (Shanghai Jiao Tong University),
Yijia Fan (Microsoft).

\vspace{0.55em}
\noindent{\sffamily\bfseries\color{argusdeep}Contributors}\par
\vspace{0.15em}
\noindent
Chuan Wen (Shanghai Jiao Tong University),
Fan Yang (Microsoft),
Hangxi Guo (The Chinese University of Hong Kong, Shenzhen),
Jiaao Wu (Tsinghua University),
Jiachen Zhang (Independent Researcher),
Junxiang Lei (Fudan University),
Mukai Li (The University of Hong Kong),
Ruize Tang (Microsoft),
Runjing Gu (Shanghai Jiao Tong University),
Shibo Hu (Peking University),
Sihan Chen (Shanghai Jiao Tong University),
Sufeng Guo (Nanjing University),
Wanbo Zhang (Fudan University),
Xian Zhang (Microsoft),
Xiaoyu Chen (Shanghai Jiao Tong University),
Xuanhe Zhou (Shanghai Jiao Tong University),
Xuyao Huang (Shanghai Jiao Tong University),
Yifei Gao (Shanghai Jiao Tong University),
Yifei Shen (Microsoft),
Yilin Chen (Donghua University),
Yuheng Wu (Shanghai Jiao Tong University),
Yuzhe Zhang (Shanghai Jiao Tong University),
Zelong Zhao (Independent Researcher),
Zhijie Deng (Shanghai Jiao Tong University).

\appendix
\section{Detailed SWE-Bench Pro Results}
\label{app:detailed-results}

The main text reports the startup-to-mature comparison. Table~\ref{tab:swebench-windows}
shows every analysis window, including the composition shift and late difficult-task period.

\begin{table}[H]
\centering
\footnotesize
\renewcommand{\arraystretch}{1.10}
\begin{tabular}{@{}llrrrr@{}}
\toprule
\rowcolor{papersand}
\textbf{Window} & \textbf{Interpretation} & \textbf{Tasks} & \textbf{Input/task} & \textbf{Active min/task} & \textbf{Token index} \\
\midrule
W1--6 & Startup & 120 & 2.95M & 8.52 & 100 \\
W7--12 & Early reuse & 140 & 2.07M & 7.28 & 70 \\
W13--18 & Composition shift & 151 & 1.47M & 10.42 & 50 \\
W19--22 & Mature & 158 & 2.33M & 7.25 & 79 \\
W23--24 & Late difficult tasks & 49 & 3.72M & 9.01 & 126 \\
\bottomrule
\end{tabular}
\caption{Task-weighted \systemname window statistics. Token index normalizes
W1--6 to 100.}
\label{tab:swebench-windows}
\end{table}

\begin{table}[H]
\centering
\footnotesize
\renewcommand{\arraystretch}{1.00}
\begin{tabular}{@{}rrrrrr@{}}
\toprule
\rowcolor{papersand}
\textbf{Wave} & \textbf{Tasks} & \textbf{Solve input/task} & \textbf{Agent s/task} & \textbf{Skills} & \textbf{Wiki} \\
\midrule
1  & 20 & 2.839M & 545.6 & 34  & 27  \\
2  & 20 & 3.263M & 642.6 & 54  & 41  \\
3  & 20 & 3.609M & 569.6 & 68  & 52  \\
4  & 20 & 2.908M & 490.7 & 89  & 64  \\
5  & 20 & 2.832M & 452.0 & 106 & 81  \\
6  & 20 & 2.246M & 366.8 & 127 & 94  \\
7  & 20 & 2.564M & 519.8 & 148 & 101 \\
8  & 20 & 2.882M & 465.8 & 168 & 115 \\
9  & 20 & 2.615M & 394.2 & 190 & 129 \\
10 & 20 & 2.204M & 377.1 & 208 & 143 \\
11 & 20 & 2.041M & 396.9 & 225 & 155 \\
12 & 40 & 1.099M & 451.5 & 260 & 171 \\
14 & 32 & 1.327M & 527.0 & 290 & 191 \\
16 & 39 & 1.564M & 623.1 & 305 & 206 \\
17 & 40 & 1.537M & 668.5 & 306 & 219 \\
18 & 40 & 1.432M & 662.8 & 333 & 237 \\
19 & 40 & 2.424M & 431.0 & 365 & 258 \\
20 & 40 & 2.129M & 400.5 & 404 & 284 \\
21 & 39 & 2.165M & 427.9 & 413 & 305 \\
22 & 39 & 2.593M & 481.3 & 434 & 323 \\
23 & 38 & 3.873M & 565.6 & 471 & 345 \\
24 & 11 & 3.186M & 455.4 & 478 & 352 \\
\bottomrule
\end{tabular}
\caption{All completed \systemname Wave summaries used in the longitudinal
analysis. Two incomplete Waves are omitted from grouped means.}
\label{tab:swebench-waves}
\end{table}

\section{Reviewer Intervention Details}

Table~\ref{tab:reviewer-mechanism} separates routing decisions from recovery
outcomes. The first two rows give the routing split; rows three through five
partition the 466 independent-Reviewer outcomes; the final two rows measure recovery
after a revision request.

\begin{table}[H]
\centering
\footnotesize
\renewcommand{\arraystretch}{1.12}
\begin{tabular}{@{}lrr@{}}
\toprule
\rowcolor{papersand}
\textbf{Reviewer mechanism event} & \textbf{Tasks} & \textbf{Interpretation} \\
\midrule
Independent Reviewer invoked & 466 & 63.7\% of 731 \\
Engineer self-review & 265 & 36.3\% of 731 \\
Accepted on first independent review & 388 & terminal \code{done} \\
Blocked on first independent review & 35 & terminal \code{blocked} \\
Independent Reviewer requested revision & 43 & at least one \code{continue} \\
Official verifier success after revision & 34 & 79.1\% of revision requests \\
Strict review-loop rescue & 22 & 51.2\% of revision requests \\
\bottomrule
\end{tabular}
\caption{Reviewer routing and recovery. Strict rescue requires an independent
Reviewer \code{continue} followed by a later Reviewer \code{done}.}
\label{tab:reviewer-mechanism}
\end{table}

\section{Representative Vertical Trace Details}

Table~\ref{tab:erdos-trace-detail} expands the role handoffs summarized in
Figure~\ref{fig:erdos-vertical-trace}. The table deliberately foregrounds Agent
authority and durable outputs rather than the mathematical statements.

\begin{table}[H]
\centering
\scriptsize
\renewcommand{\arraystretch}{1.10}
\resizebox{\linewidth}{!}{%
\begin{tabular}{@{}lp{2.55cm}p{2.55cm}p{2.75cm}p{3.7cm}@{}}
\toprule
\rowcolor{papersand}
\textbf{Phase} & \textbf{Manager / Planner} & \textbf{Engineer} & \textbf{Reviewer} & \textbf{Durable output} \\
\midrule
Recover and scope & Manager restores the campaign; Planner screens candidate problems. & Grounds sources and creates deterministic checks. & Verifies problem fidelity and accepts the scoped objective. & Persistent objective, selected problem, initial claim record. \\
Test a route & Planner chooses the cheapest decisive falsification mission. & Runs the witness verifier and records the failed reduction. & Accepts the negative result and fixes its scope. & Certified dead branch and revised next action. \\
Produce proof package & Planner requires theorem, proof, and verification artifacts. & Writes proof, validation record, lemma graph, and checkpoint. & Returns \code{continue} when checks are incomplete, then rechecks. & Reviewer-accepted bounded result package. \\
Strengthen & Planner requires later work to consume and advance retained state. & Improves one result and proves bridge/obstruction lemmas. & Checks correctness, strict progress, and claim boundaries. & Updated accepted state plus named remaining blockers. \\
Continue & Manager advances or rolls back from the verdict; Planner selects the next blocker. & Starts the next mission from retained artifacts. & Updates the frontier status and next blockers. & Inspectable research trace and reusable cross-session state. \\
\bottomrule
\end{tabular}%
}
\caption{Role-resolved phases in the reported Erd\H{o}s--Gy\'arf\'as campaign.
The table follows how each role changes the shared research state.}
\label{tab:erdos-trace-detail}
\end{table}

The supplementary trace bundle contains the role/claim table, calculation map,
editable figure source, and provenance. The source campaign retains the complete
artifacts and event history.

\begin{paperbox}{Representative proof artifact (abridged)}
\small
For fixed roots $x,y$, let $\mathcal{P}_4(x,y)$ be the length-$4$ paths from
$x$ to $y$, and let $\mathcal{F}$ be their internal three-vertex sets. In a
graph with no $C_8$, any two members of $\mathcal{F}$ intersect: two paths with
disjoint internal vertices would join to form an $8$-cycle. A separate
no-$C_4$ case analysis shows that any fixed pair of internal vertices occurs in
at most three path sets.

Assume that the path family has no common internal blocker. Choose
$E=\{a,b,c\}\in\mathcal{F}$. For each $t\in E$, some member $F_t$ omits $t$.
Every set containing $t$ must meet the three vertices of $F_t$; the pair-occurrence
bound therefore gives $d(t)\leq 9$. Hence
\[
  d(a)+d(b)+d(c)\leq 27.
\]
Every member of $\mathcal{F}$ meets $E$, while $E$ itself contributes three
incidences rather than one, so
\[
  |\mathcal{F}|+2\leq d(a)+d(b)+d(c)\leq 27.
\]
Thus $|\mathcal{P}_4(x,y)|=|\mathcal{F}|\leq25$: the paths share a common
internal blocker or there are at most $25$ of them. This excerpt illustrates the
kind of proof artifact produced and reviewed by the workflow; the complete case
analysis and computational checks remain in the source bundle.
\end{paperbox}

\section{Process-Theory Calculations}
\label{app:formula-instantiation}

Table~\ref{tab:formula-instantiation} records the complete substitutions behind the
process quantities summarized in the main text. The mathematical window contains
332,274 reasoning-output Tokens over 9,620.5 active seconds. The strict and
inclusive density values use the two attribution policies defined in
Equation~\ref{eq:operational-efficiency}. The process-compression row is a separate
campaign snapshot.

\begin{table}[H]
\centering
\scriptsize
\renewcommand{\arraystretch}{1.13}
\begin{tabularx}{\linewidth}{@{}p{2.25cm}p{4.3cm}p{2.15cm}X@{}}
\toprule
\rowcolor{papersand}
\textbf{Theory quantity} & \textbf{Data substitution} & \textbf{Observed value} & \textbf{Interpretation} \\
\midrule
Strict $\widehat{\rho}_I$ & $332{,}274/9{,}620.5\times0.5598\times0.5556\times0.5538$ & 5.95 effective Token-eq./s & Direct reasoning, action, and verification only. \\
Inclusive $\widehat{\rho}_I$ & $332{,}274/9{,}620.5\times0.9512\times0.9444\times0.9043$ & 28.06 effective Token-eq./s & Includes successful coordination and state-maintenance work. \\
Reviewer correction & $34/43$ verifier recoveries; $22/43$ strict rescues & 79.1\%; 51.2\% & Post-revision outcomes on SWE-Bench Pro. \\
Reuse quantity $G_L$ & $(2.95-2.33)$M Tokens/task; $(8.52-7.25)$ min/task & 0.62M Tokens and 1.27 min per task & Startup--mature difference in the longitudinal run. \\
Immediate $\widehat{\mathcal{Y}}_{\mathrm{PRI}}$ & $\lambda_g=0$; $6/332{,}274\times10^6$ & 18.1 accepted deltas/M reasoning Tokens & Six accepted theorem-frontier updates in Rounds 12--17. \\
Process compression & $384{,}370{,}463/3{,}240$ bytes & 118,633$\times$ event-history/checkpoint ratio & Event history relative to the active working checkpoint. \\
\bottomrule
\end{tabularx}
\caption{Detailed calculation of the process-theory quantities. Values retain
their task-specific units; the main text reports the corresponding system-level
patterns.}
\label{tab:formula-instantiation}
\end{table}

\section{Upstream Kernel Adoption}
\label{app:upstream-kernel}

The kernel-optimization work produced an externally reviewed upstream
contribution to Flash Linear Attention. In
\href{https://github.com/fla-org/flash-linear-attention/pull/1045}
{FLA PR \#1045}, \systemname implemented and optimized an opt-in TileLang
\code{RWKV6} dense-\code{bf16}, \code{D=64} forward-intra kernel. The submitted
H100 evidence reports a forward latency reduction from 0.199 ms to 0.168 ms
(1.18$\times$) and a forward-plus-backward reduction from 0.900 ms to 0.747 ms
(1.21$\times$), with 13 correctness-gate passes and 14 repository tests.

An FLA collaborator affiliated with Moonshot AI reviewed the generated CUDA,
identified a long-sequence numerical-stability issue, and requested block-local
exponent centering. After \systemname applied the fix and reran the verification
suite, the contribution was merged into \code{fla-org:main} on July 20, 2026 as
\href{https://github.com/fla-org/flash-linear-attention/commit/c70f11c5530142450525549cc96d13d9f5165f69}
{commit \code{c70f11c}}. This is evidence of human-reviewed upstream adoption,
not external scientific validation of the broader runtime.

\section{Notation}

\begin{table}[H]
\centering
\footnotesize
\begin{tabularx}{0.88\linewidth}{@{}p{2.6cm} X@{}}
\toprule
\rowcolor{papersand}
\textbf{Symbol} & \textbf{Meaning} \\
\midrule
$o,y_0,f,B$ & Objective, initial artifact, evaluator, and resource budget \\
$K_t=(\iota,o_t,c_t,v_t)$ & Standing intent, objective, constraints, and verification criteria \\
$X_t$ & User-visible clarifications, priorities, and unresolved questions \\
$K'_t,u_t$ & Proposed contract and required admission authority \\
$\Gamma_t$ & Round-level operational records before trajectory projection \\
$\tau_t$ & Mission trajectory over one or more execution/review rounds \\
$H_t$ & Persistent runtime state \\
$E_t$ & Results admitted after mission $t$ \\
$Q_t$ & Persistent task and evaluation definitions \\
$U$ & Partial persistent-state update operator \\
$\operatorname{ManagerAdmit}$ & Analytical projection of evidence- and authority-gated contract refinement \\
$\theta_t$ & Underlying model parameters, fixed in this report \\
$C_t$ & Effective retained capability set \\
$\phi(\tau_t)$ & Role-resolved mission trace $(m_t,p_t,x_t,r_t,\Delta H_t)$ \\
$\rho_I(T)$ & Dense-intelligence density over horizon $T$ \\
$\eta_r,\eta_a,\eta_v$ & Measured reasoning, action, and verification efficiency factors \\
$\mathcal{R}_q(X)$ & Minimum downstream decision risk when observing $X$ \\
$\psi_b^\star$ & Decision-useful process compression under context budget $b$ \\
$G_L(\Delta H_t)$ & Discounted counterfactual reuse value over $L$ future tasks \\
$\mathcal{Y}_{\mathrm{PRI}}(W)$ & Verified reusable process-intelligence yield per Token \\
$\alpha,\beta$ & Reviewer sensitivity and false-acceptance rate \\
$V_R(T)$ & Conceptual accumulated research value \\
$R_{\mathrm{tok}}$ & Aggregate \systemname-to-Direct-Copilot Token ratio \\
$\bar I_w^{\mathrm{solve}}$ & Mean solve input tokens in Wave $w$ \\
$\bar T_w^{\mathrm{agent}}$ & Mean active workflow time in Wave $w$ \\
\bottomrule
\end{tabularx}
\caption{Notation used in the report.}
\end{table}

Machine-readable aggregates and editable figure sources are released with the
report. They are kept outside the narrative so that implementation metadata does
not obscure the scientific results.

\clearpage
\bibliographystyle{plainnat}
\bibliography{references}

@article{yao2023react,
  author       = {Yao, Shunyu and others},
  title        = {{ReAct}: Synergizing Reasoning and Acting in Language Models},
  journal      = {arXiv preprint arXiv:2210.03629},
  year         = {2023},
  eprint       = {2210.03629},
  archivePrefix= {arXiv},
  primaryClass = {cs.CL},
  url          = {https://arxiv.org/abs/2210.03629}
}

@article{shinn2023reflexion,
  author       = {Shinn, Noah and others},
  title        = {Reflexion: Language Agents with Verbal Reinforcement Learning},
  journal      = {arXiv preprint arXiv:2303.11366},
  year         = {2023},
  eprint       = {2303.11366},
  archivePrefix= {arXiv},
  primaryClass = {cs.AI},
  url          = {https://arxiv.org/abs/2303.11366}
}

@article{wang2023voyager,
  author       = {Wang, Guanzhi and others},
  title        = {Voyager: An Open-Ended Embodied Agent with Large Language Models},
  journal      = {arXiv preprint arXiv:2305.16291},
  year         = {2023},
  eprint       = {2305.16291},
  archivePrefix= {arXiv},
  primaryClass = {cs.AI},
  url          = {https://arxiv.org/abs/2305.16291}
}

@article{schick2023toolformer,
  author       = {Schick, Timo and Dwivedi-Yu, Jane and Dess{\`i}, Roberto and others},
  title        = {Toolformer: Language Models Can Teach Themselves to Use Tools},
  journal      = {arXiv preprint arXiv:2302.04761},
  year         = {2023},
  eprint       = {2302.04761},
  archivePrefix= {arXiv},
  primaryClass = {cs.CL},
  url          = {https://arxiv.org/abs/2302.04761}
}

@article{park2023generative,
  author       = {Park, Joon Sung and O'Brien, Joseph C. and Cai, Carrie J. and others},
  title        = {Generative Agents: Interactive Simulacra of Human Behavior},
  journal      = {arXiv preprint arXiv:2304.03442},
  year         = {2023},
  eprint       = {2304.03442},
  archivePrefix= {arXiv},
  primaryClass = {cs.HC},
  url          = {https://arxiv.org/abs/2304.03442}
}

@article{packer2023memgpt,
  author       = {Packer, Charles and Wooders, Sarah and Lin, Kevin and others},
  title        = {{MemGPT}: Towards {LLMs} as Operating Systems},
  journal      = {arXiv preprint arXiv:2310.08560},
  year         = {2023},
  eprint       = {2310.08560},
  archivePrefix= {arXiv},
  primaryClass = {cs.AI},
  url          = {https://arxiv.org/abs/2310.08560}
}

@article{wang2024workflowmemory,
  author       = {Wang, Zora Zhiruo and Mao, Jiayuan and Fried, Daniel and Neubig, Graham},
  title        = {Agent Workflow Memory},
  journal      = {arXiv preprint arXiv:2409.07429},
  year         = {2024},
  eprint       = {2409.07429},
  archivePrefix= {arXiv},
  primaryClass = {cs.CL},
  url          = {https://arxiv.org/abs/2409.07429}
}

@article{lightman2023verify,
  author       = {Lightman, Hunter and Kosaraju, Vineet and Burda, Yura and
                  Edwards, Harri and Baker, Bowen and Lee, Teddy and Leike, Jan and
                  Schulman, John and Sutskever, Ilya and Cobbe, Karl},
  title        = {Let's Verify Step by Step},
  journal      = {arXiv preprint arXiv:2305.20050},
  year         = {2023},
  eprint       = {2305.20050},
  archivePrefix= {arXiv},
  primaryClass = {cs.LG},
  url          = {https://arxiv.org/abs/2305.20050}
}

@article{snell2024testtime,
  author       = {Snell, Charlie and Lee, Jaehoon and Xu, Kelvin and Kumar, Aviral},
  title        = {Scaling {LLM} Test-Time Compute Optimally Can Be More Effective
                  than Scaling Model Parameters},
  journal      = {arXiv preprint arXiv:2408.03314},
  year         = {2024},
  eprint       = {2408.03314},
  archivePrefix= {arXiv},
  primaryClass = {cs.LG},
  url          = {https://arxiv.org/abs/2408.03314}
}

@article{blackwell1953equivalent,
  author       = {Blackwell, David},
  title        = {Equivalent Comparisons of Experiments},
  journal      = {The Annals of Mathematical Statistics},
  volume       = {24},
  number       = {2},
  pages        = {265--272},
  year         = {1953},
  doi          = {10.1214/aoms/1177729032},
  url          = {https://doi.org/10.1214/aoms/1177729032}
}

@article{baek2024researchagent,
  author       = {Baek, Jinheon and Jauhar, Sujay Kumar and Cucerzan, Silviu and Hwang, Sung Ju},
  title        = {{ResearchAgent}: Iterative Research Idea Generation over Scientific Literature with Large Language Models},
  journal      = {arXiv preprint arXiv:2404.07738},
  year         = {2024},
  eprint       = {2404.07738},
  archivePrefix= {arXiv},
  primaryClass = {cs.CL},
  url          = {https://arxiv.org/abs/2404.07738}
}

@article{ghafarollahi2024sciagents,
  author       = {Ghafarollahi, Alireza and Buehler, Markus J.},
  title        = {{SciAgents}: Automating Scientific Discovery through Multi-Agent Intelligent Graph Reasoning},
  journal      = {arXiv preprint arXiv:2409.05556},
  year         = {2024},
  eprint       = {2409.05556},
  archivePrefix= {arXiv},
  primaryClass = {cs.AI},
  url          = {https://arxiv.org/abs/2409.05556}
}

@inproceedings{weng2025cycleresearcher,
  author       = {Weng, Yixuan and Zhu, Minjun and Bao, Guangsheng and Zhang, Hongbo and Wang, Jindong and Zhang, Yue and Yang, Linyi},
  title        = {{CycleResearcher}: Improving Automated Research via Automated Review},
  booktitle    = {International Conference on Learning Representations},
  year         = {2025},
  url          = {https://openreview.net/forum?id=bjcsVLoHYs}
}

@article{gottweis2025coscientist,
  author       = {Gottweis, Juraj and Weng, Wei-Hung and Daryin, Alexander and others},
  title        = {Accelerating Scientific Discovery with Co-Scientist},
  journal      = {arXiv preprint arXiv:2502.18864},
  year         = {2025},
  eprint       = {2502.18864},
  archivePrefix= {arXiv},
  primaryClass = {cs.AI},
  url          = {https://arxiv.org/abs/2502.18864}
}

@article{schmidgall2025agentrxiv,
  author       = {Schmidgall, Samuel and Moor, Michael},
  title        = {{AgentRxiv}: Towards Collaborative Autonomous Research},
  journal      = {arXiv preprint arXiv:2503.18102},
  year         = {2025},
  eprint       = {2503.18102},
  archivePrefix= {arXiv},
  primaryClass = {cs.AI},
  url          = {https://arxiv.org/abs/2503.18102}
}

@article{qian2026autosci,
  author       = {Qian, Weitong and Xu, Beicheng and Xie, Zhongao and others},
  title        = {{AutoSci}: A Memory-Centric Agentic System for the Full Scientific Research Lifecycle},
  journal      = {arXiv preprint arXiv:2605.31468},
  year         = {2026},
  eprint       = {2605.31468},
  archivePrefix= {arXiv},
  primaryClass = {cs.AI},
  url          = {https://arxiv.org/abs/2605.31468}
}

@article{tang2026fars,
  author       = {Tang, Qiong and Sun, Tianxiang and Hu, Xiangkun and others},
  title        = {{FARS}: A Fully Automated Research System Deployed at Scale},
  journal      = {arXiv preprint arXiv:2606.31651},
  year         = {2026},
  eprint       = {2606.31651},
  archivePrefix= {arXiv},
  primaryClass = {cs.AI},
  url          = {https://arxiv.org/abs/2606.31651}
}

@article{mitra2024agentinstruct,
  author       = {Mitra, Arindam and Del Corro, Luciano and Zheng, Guoqing and
                  Mahajan, Shweti and Rouhana, Dany and Codas, Andres and others},
  title        = {{AgentInstruct}: Toward Generative Teaching with Agentic Flows},
  journal      = {arXiv preprint arXiv:2407.03502},
  year         = {2024},
  eprint       = {2407.03502},
  archivePrefix= {arXiv},
  primaryClass = {cs.CL},
  url          = {https://arxiv.org/abs/2407.03502}
}

@inproceedings{chen2024agentflan,
  author       = {Chen, Zehui and Liu, Kuikun and Wang, Qiuchen and Zhang, Wenwei
                  and Liu, Jiangning and Lin, Dahua and Chen, Kai and Zhao, Feng},
  title        = {{Agent-FLAN}: Designing Data and Methods of Effective Agent
                  Tuning for Large Language Models},
  booktitle    = {Findings of the Association for Computational Linguistics: ACL 2024},
  pages        = {9354--9366},
  publisher    = {Association for Computational Linguistics},
  year         = {2024},
  doi          = {10.18653/v1/2024.findings-acl.557},
  url          = {https://aclanthology.org/2024.findings-acl.557/}
}

@article{luo2025agentlightning,
  author       = {Luo, Xufang and Zhang, Yuge and He, Zhiyuan and Wang, Zilong and
                  Zhao, Siyun and Li, Dongsheng and Qiu, Luna K. and Yang, Yuqing},
  title        = {Agent Lightning: Train Any {AI} Agents with Reinforcement Learning},
  journal      = {arXiv preprint arXiv:2508.03680},
  year         = {2025},
  eprint       = {2508.03680},
  archivePrefix= {arXiv},
  primaryClass = {cs.LG},
  url          = {https://arxiv.org/abs/2508.03680}
}

@article{xu2025amem,
  author       = {Xu, Wujiang and Liang, Zujie and Mei, Kai and others},
  title        = {{A-MEM}: Agentic Memory for {LLM} Agents},
  journal      = {arXiv preprint arXiv:2502.12110},
  year         = {2025},
  eprint       = {2502.12110},
  archivePrefix= {arXiv},
  primaryClass = {cs.AI},
  url          = {https://arxiv.org/abs/2502.12110}
}

@article{yang2024sweagent,
  author       = {Yang, John and others},
  title        = {{SWE-agent}: Agent-Computer Interfaces Enable Automated Software Engineering},
  journal      = {arXiv preprint arXiv:2405.15793},
  year         = {2024},
  eprint       = {2405.15793},
  archivePrefix= {arXiv},
  primaryClass = {cs.SE},
  url          = {https://arxiv.org/abs/2405.15793}
}

@article{wang2024openhands,
  author       = {Wang, Xingyao and Li, Boxuan and Song, Yufan and others},
  title        = {{OpenHands}: An Open Platform for {AI} Software Developers as Generalist Agents},
  journal      = {arXiv preprint arXiv:2407.16741},
  year         = {2024},
  eprint       = {2407.16741},
  archivePrefix= {arXiv},
  primaryClass = {cs.SE},
  url          = {https://arxiv.org/abs/2407.16741}
}

@article{jimenez2024swebench,
  author       = {Jimenez, Carlos E. and Yang, John and Wettig, Alexander and others},
  title        = {{SWE-bench}: Can Language Models Resolve Real-World {GitHub} Issues?},
  journal      = {arXiv preprint arXiv:2310.06770},
  year         = {2024},
  eprint       = {2310.06770},
  archivePrefix= {arXiv},
  primaryClass = {cs.SE},
  url          = {https://arxiv.org/abs/2310.06770}
}

@article{deng2025swebenchpro,
  author       = {Xiang Deng and Jeff Da and Edwin Pan and Yannis Yiming He and
                  Charles Ide and Kanak Garg and Niklas Lauffer and Andrew Park and
                  Nitin Pasari and Chetan Rane and Karmini Sampath and Maya Krishnan and
                  Srivatsa Kundurthy and Sean Hendryx and Zifan Wang and
                  Vijay Bharadwaj and Jeff Holm and Raja Aluri and
                  Chen Bo Calvin Zhang and Noah Jacobson and Bing Liu and Brad Kenstler},
  title        = {{SWE-Bench Pro}: Can {AI} Agents Solve Long-Horizon Software Engineering Tasks?},
  journal      = {arXiv preprint arXiv:2509.16941},
  year         = {2025},
  eprint       = {2509.16941},
  archivePrefix= {arXiv},
  primaryClass = {cs.SE},
  url          = {https://arxiv.org/abs/2509.16941}
}

@article{liu2023agentbench,
  author       = {Liu, Xiao and Yu, Hao and Zhang, Hanchen and others},
  title        = {{AgentBench}: Evaluating {LLMs} as Agents},
  journal      = {arXiv preprint arXiv:2308.03688},
  year         = {2023},
  eprint       = {2308.03688},
  archivePrefix= {arXiv},
  primaryClass = {cs.AI},
  url          = {https://arxiv.org/abs/2308.03688}
}

@article{ma2024agentboard,
  author       = {Ma, Chang and Zhang, Junlei and Zhu, Zhihao and others},
  title        = {{AgentBoard}: An Analytical Evaluation Board of Multi-turn {LLM} Agents},
  journal      = {arXiv preprint arXiv:2401.13178},
  year         = {2024},
  eprint       = {2401.13178},
  archivePrefix= {arXiv},
  primaryClass = {cs.CL},
  url          = {https://arxiv.org/abs/2401.13178}
}

@article{kwa2025longtasks,
  author       = {Kwa, Thomas and West, Ben and Becker, Joel and others},
  title        = {Measuring {AI} Ability to Complete Long Software Tasks},
  journal      = {arXiv preprint arXiv:2503.14499},
  year         = {2025},
  eprint       = {2503.14499},
  archivePrefix= {arXiv},
  primaryClass = {cs.AI},
  url          = {https://arxiv.org/abs/2503.14499}
}

@article{lu2024aiscientist,
  author       = {Lu, Chris and others},
  title        = {The {AI} Scientist: Towards Fully Automated Open-Ended Scientific Discovery},
  journal      = {arXiv preprint arXiv:2408.06292},
  year         = {2024},
  eprint       = {2408.06292},
  archivePrefix= {arXiv},
  primaryClass = {cs.AI},
  url          = {https://arxiv.org/abs/2408.06292}
}

@article{schmidgall2025agentlab,
  author       = {Schmidgall, Samuel and Su, Yusheng and Wang, Ze and others},
  title        = {Agent Laboratory: Using {LLM} Agents as Research Assistants},
  journal      = {arXiv preprint arXiv:2501.04227},
  year         = {2025},
  eprint       = {2501.04227},
  archivePrefix= {arXiv},
  primaryClass = {cs.CL},
  url          = {https://arxiv.org/abs/2501.04227}
}

@article{jiang2025aide,
  author       = {Jiang, Zhengyao and Schmidt, Dominik and Srikanth, Dhruv and others},
  title        = {{AIDE}: {AI}-Driven Exploration in the Space of Code},
  journal      = {arXiv preprint arXiv:2502.13138},
  year         = {2025},
  eprint       = {2502.13138},
  archivePrefix= {arXiv},
  primaryClass = {cs.LG},
  url          = {https://arxiv.org/abs/2502.13138}
}

@article{yamada2025aiscientistv2,
  author       = {Yamada, Yutaro and Lange, Robert Tjarko and Lu, Cong and others},
  title        = {The {AI} Scientist-v2: Workshop-Level Automated Scientific Discovery via Agentic Tree Search},
  journal      = {arXiv preprint arXiv:2504.08066},
  year         = {2025},
  eprint       = {2504.08066},
  archivePrefix= {arXiv},
  primaryClass = {cs.AI},
  url          = {https://arxiv.org/abs/2504.08066}
}

@article{romera2024funsearch,
  author       = {Romera-Paredes, Bernardino and Barekatain, Mohammadamin and Novikov, Alexander and others},
  title        = {Mathematical Discoveries from Program Search with Large Language Models},
  journal      = {Nature},
  volume       = {625},
  pages        = {468--475},
  year         = {2024},
  doi          = {10.1038/s41586-023-06924-6},
  url          = {https://doi.org/10.1038/s41586-023-06924-6}
}

@article{novikov2025alphaevolve,
  author       = {Novikov, Alexander and V{\~u}, Ng{\^a}n and Eisenberger, Marvin and others},
  title        = {{AlphaEvolve}: A Coding Agent for Scientific and Algorithmic Discovery},
  journal      = {arXiv preprint arXiv:2506.13131},
  year         = {2025},
  eprint       = {2506.13131},
  archivePrefix= {arXiv},
  primaryClass = {cs.AI},
  url          = {https://arxiv.org/abs/2506.13131}
}

@article{jin2026arbor,
  author       = {Jin, Jiajie and Hu, Yuyang and Qiu, Kai and Dai, Qi and Luo, Chong and others},
  title        = {Toward Generalist Autonomous Research via Hypothesis-Tree Refinement},
  journal      = {arXiv preprint arXiv:2606.11926},
  year         = {2026},
  eprint       = {2606.11926},
  archivePrefix= {arXiv},
  primaryClass = {cs.CL},
  url          = {https://arxiv.org/abs/2606.11926}
}

@article{chan2024mlebench,
  author       = {Chan, Jun Shern and Chowdhury, Neil and Jaffe, Oliver and others},
  title        = {{MLE-bench}: Evaluating Machine Learning Agents on Machine Learning Engineering},
  journal      = {arXiv preprint arXiv:2410.07095},
  year         = {2024},
  eprint       = {2410.07095},
  archivePrefix= {arXiv},
  primaryClass = {cs.CL},
  url          = {https://arxiv.org/abs/2410.07095}
}

@article{wijk2024rebench,
  author       = {Wijk, Hjalmar and Lin, Tao and Becker, Joel and others},
  title        = {{RE-Bench}: Evaluating Frontier {AI} {R\&D} Capabilities of Language Model Agents against Human Experts},
  journal      = {arXiv preprint arXiv:2411.15114},
  year         = {2024},
  eprint       = {2411.15114},
  archivePrefix= {arXiv},
  primaryClass = {cs.LG},
  url          = {https://arxiv.org/abs/2411.15114}
}

@article{wang2026aarri,
  author       = {Wang, Jiayu and Lv, Weijiang and Fu, Bowen and others},
  title        = {Act As a Real Researcher: A Suite of Benchmarks Evaluating Frontier {LLMs} and Agentic Harnesses in Research Lifecycle},
  journal      = {arXiv preprint arXiv:2606.07462},
  year         = {2026},
  eprint       = {2606.07462},
  archivePrefix= {arXiv},
  primaryClass = {cs.AI},
  url          = {https://arxiv.org/abs/2606.07462}
}

@article{lupidi2026airsbench,
  author       = {Lupidi, Alisia and Gauri, Bhavul and Foster, Thomas Simon and others},
  title        = {{AIRS-Bench}: A Suite of Tasks for Frontier {AI} Research Science Agents},
  journal      = {arXiv preprint arXiv:2602.06855},
  year         = {2026},
  eprint       = {2602.06855},
  archivePrefix= {arXiv},
  primaryClass = {cs.AI},
  url          = {https://arxiv.org/abs/2602.06855}
}

@article{lin2026solexecbench,
  author       = {Lin, Edward and Modi, Sahil and Hari, Siva Kumar Sastry and others},
  title        = {{SOL-ExecBench}: Speed-of-Light Benchmarking for Real-World {GPU} Kernels Against Hardware Limits},
  journal      = {arXiv preprint arXiv:2603.19173},
  year         = {2026},
  eprint       = {2603.19173},
  archivePrefix= {arXiv},
  primaryClass = {cs.DC},
  url          = {https://arxiv.org/abs/2603.19173}
}

@misc{karpathy_nanochat,
  author       = {Karpathy, Andrej},
  title        = {{nanochat}: The best {ChatGPT} that \$100 can buy},
  year         = {2025},
  howpublished = {GitHub repository},
  url          = {https://github.com/karpathy/nanochat}
}

@misc{karpathy_autoresearch,
  author       = {Karpathy, Andrej},
  title        = {autoresearch: {AI} Agents Running Research on Single-{GPU} nanochat Training Automatically},
  year         = {2026},
  howpublished = {GitHub repository},
  url          = {https://github.com/karpathy/autoresearch}
}

@misc{jordan_nanogptbench,
  author       = {Jordan, Keller and contributors},
  title        = {{Modded-NanoGPT}: The NanoGPT Speedrun},
  year         = {2024},
  howpublished = {GitHub repository},
  url          = {https://github.com/KellerJordan/modded-nanogpt}
}

@misc{argus_site,
  author       = {{Argus Team}},
  title        = {Argus: A Self-Evolving Research Agent},
  year         = {2026},
  howpublished = {Project website},
  url          = {https://argusbot.cn/}
}

\end{document}